\documentclass[letterpaper]{article} 
\usepackage[preprint]{aaai2027}  
\usepackage[hyphens]{url}  
\usepackage{graphicx} 
\usepackage{amsmath}
\usepackage{amssymb}

\usepackage[table]{xcolor}
\usepackage{amsmath,amssymb}
\usepackage[caption=false,font=footnotesize]{subfig}

\usepackage{booktabs}
\usepackage{multirow}
\usepackage{natbib}  
\usepackage{caption} 
\usepackage{algorithm}
\usepackage{algorithmic}

\usepackage{newfloat}
\usepackage{listings}
\DeclareCaptionStyle{ruled}{labelfont=normalfont,labelsep=colon,strut=off} 
\floatstyle{ruled}
\newfloat{listing}{tb}{lst}{}
\floatname{listing}{Listing}

\usepackage{booktabs}

\makeatletter
\newcommand{\multiplecorrespondingauthors}{\global\aaai@corrmultitrue}
\makeatother

\title{MomADv2: Reliable Temporal Memory for End-to-End Autonomous Driving}
\author{
    \multiplecorrespondingauthors
    Ziying Song\textsuperscript{\rm 1},
    Shengkai Zhang\textsuperscript{\rm 2},
    Lin Liu\textsuperscript{\rm 2},
    Peiliang Wu\textsuperscript{\rm 1},
    Lei Yang\textsuperscript{\rm 3}\corresponding,\\
    Dongyang Xu\textsuperscript{\rm 4},
    Bin Sun\textsuperscript{\rm 5},
    Li Wang\textsuperscript{\rm 6}\corresponding,
    Shaoqing Xu\textsuperscript{\rm 7},
    Caiyan Jia\textsuperscript{\rm 2},
    Yadan Luo\textsuperscript{\rm 8}
}
\affiliations{
    \textsuperscript{\rm 1}School of Artificial Intelligence (School of Software), Yanshan University\\
    \textsuperscript{\rm 2}Beijing Jiaotong University\\
    \textsuperscript{\rm 3}Nanyang Technological University\\
    \textsuperscript{\rm 4}Tsinghua University\\
    \textsuperscript{\rm 5}China Automotive Technology and Research Center Co., Ltd.\\
    \textsuperscript{\rm 6}School of Mechanical Engineering, Beijing Institute of Technology\\
    \textsuperscript{\rm 7}University of Macau\\
    \textsuperscript{\rm 8}The University of Queensland
}

\begin{document}

\maketitle

\begin{abstract}
Long-horizon planning is critical for safe autonomous driving in complex scenarios. Existing methods improve planning continuity with temporal memory, but such memory may become invalid and mislead decisions when the driving command changes. Thus, selectively leveraging useful history while suppressing command-inconsistent memory remains a key challenge.
To address this issue, we propose \textbf{MomADv2}, a reliable state-space memory framework for long-horizon end-to-end autonomous driving. At its core, MomADv2 introduces a \textbf{Selective State-Space Planning Memory Query Module}, which filters historical planning queries based on temporal continuity and command consistency, selects planning modes relevant to the current command, and models the evolution of planning intentions through a selective state-space mechanism.
To further alleviate local trajectory deviations and error accumulation in long-horizon planning, we design a \textbf{Flow-Matching Trajectory Residual Refiner}. It learns a continuous residual correction field from the refined planning output to the expert trajectory, enabling fine-grained trajectory refinement while preserving the stability of anchor-based planning.
Extensive experiments on closed-loop NAVSIM and Bench2Drive, as well as open-loop nuScenes, demonstrate that MomADv2 improves long-horizon planning consistency and reduces the average collision rate by 15.6\% over MomAD under 6-second planning.

\end{abstract}

\section{Introduction}
\begin{figure*}[!t]
\includegraphics[width=1\linewidth]{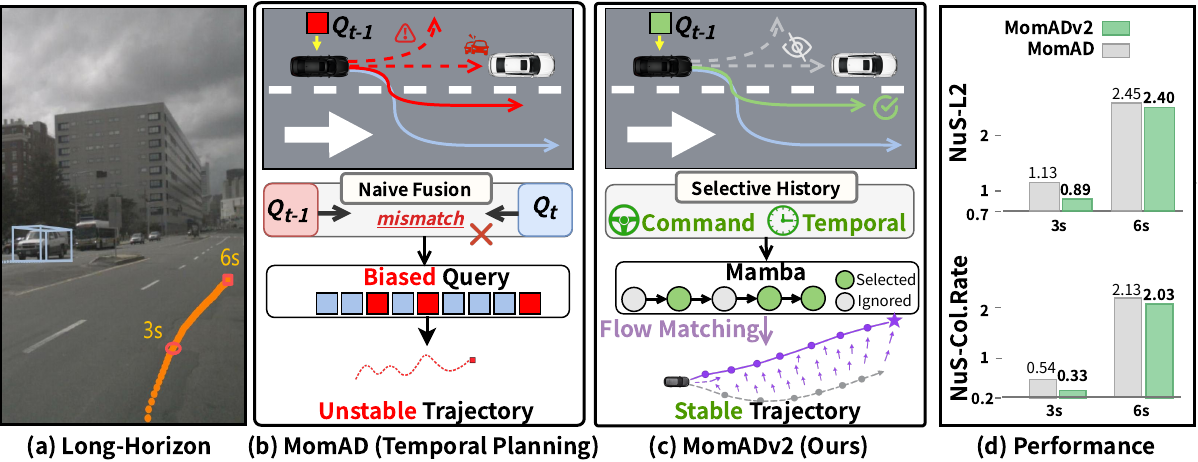}
\caption[ ]{\textbf{Motivation of MomADv2.}
(a) Long-horizon planning requires consistent intentions across planning cycles.
(b) Existing methods, represented by MomAD~\cite{momad,jia2025drivetransformer,zhang2025bridging}, indiscriminately reuse history, introducing new interference.
(c) MomADv2 selectively preserves reliable states and refines local trajectory deviations via flow matching.
(d) This reliable memory inheritance enables more accurate and safer long-horizon planning.
}

\label{fig:motivation}
\end{figure*}
End-to-end autonomous driving (E2E-AD) has recently attracted increasing attention due to its ability to integrate perception, prediction, and planning into a unified learnable framework~\cite{uniad,jiang2023vad}. Instead of separately optimizing each driving component, recent methods directly learn ego planning from sensor observations, surrounding agents, map elements, and high-level navigation commands~\cite{jia2023driveadapter,jia2025drivetransformer}. With the development of query-based perception, motion prediction, sparse scene representation, world modeling, and anchor-based planning, end-to-end driving systems have achieved promising performance in complex urban scenarios~\cite{sun2024sparsedrive,momad,xing2025goalflow}.

Long-horizon planning provides autonomous driving systems with a stronger ability to anticipate future behaviors, enabling them to proactively perceive potential risks in complex traffic interactions and generate smoother, more stable, and safer ego trajectories. A fundamental requirement of long-horizon planning is temporal memory~\cite{uniad,zhang2025bridging,jiang2023vad,momad,su2026drivemamba}. Long-horizon planning requires temporally consistent planning states across consecutive frames. MomAD~\cite{momad} shows that preserving planning momentum through historical planning queries can reduce mode instability and trajectory jitter. Therefore, effective temporal memory is essential for stable ego trajectory generation.

Existing temporal E2E-AD methods exploit history through spatial-temporal modeling, structured planning queries, task memory, or state-space representations~\cite{zhang2025bridging,momad,jia2025drivetransformer,su2026drivemamba}. 
While temporal memory improves planning continuity, it is not always reliable. 
Under dynamic scene changes, temporal discontinuities, or command shifts, historical information may become inconsistent with the current intention and mislead planning. 
Therefore, long-horizon planning should not blindly reuse history, but selectively retain reliable memory while suppressing invalid or command-inconsistent interference.


In this work, we propose \textbf{MomADv2}, a reliable state-space memory framework for long-horizon E2E-AD that selectively retains reliable history, rather than fully trusting historical information as MomAD does. Instead of blindly reusing historical planning states, MomADv2 introduces a \textbf{Selective State-Space Planning Memory Query Module (SSM-Q)} to filter historical planning queries based on temporal continuity and command consistency, select command-relevant planning modes, and model the temporal evolution of planning intentions. To further mitigate local deviations and error accumulation, we design a \textbf{Flow-Matching Trajectory Residual Refiner (FM-Ref)}, which learns a continuous residual correction field from the refined planning output to the expert trajectory, enabling fine-grained trajectory refinement while preserving the stability of anchor-based planning. Extensive experiments on NAVSIM, Bench2Drive, and nuScenes show that MomADv2 improves long-horizon planning consistency and trajectory accuracy, reducing the average 6-second collision rate by 15.6\% over MomAD \cite{momad}.
The main contributions of this work are summarized as follows:
\begin{itemize}
\item We propose \textbf{MomADv2}, a reliable state-space memory framework for long-horizon E2E-AD, which selectively leverages useful historical information while suppressing invalid memory interference.

\item We design two core modules: the \textbf{Selective State-Space Planning Memory Query Module} for reliable historical query filtering and intention modeling, and the \textbf{Flow-Matching Trajectory Residual Refiner} for fine-grained trajectory correction.

\item Extensive experiments on the closed-loop NAVSIM and Bench2Drive benchmarks, as well as the open-loop nuScenes dataset, demonstrate that MomADv2 effectively improves long-horizon planning consistency and trajectory accuracy.

\end{itemize}

\begin{figure*}[t]
\centering
\includegraphics[width=1\textwidth]{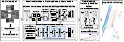}
\caption[Overview of the proposed MomADv2 framework]{Overview of \textbf{MomADv2}. Multi-view images are encoded into map and detection representations to generate initial ego planning queries. A \textbf{Selective State-Space Planning Memory Query} module retains valid historical queries based on scene, token, command, and buffer consistency. A \textbf{Flow-Matching Trajectory Residual Refiner} then predicts a conditional residual velocity field toward the expert trajectory, producing smooth, accurate, and temporally consistent long-horizon planning.}

\label{fig:framework}
\end{figure*}

\begin{figure}[t]
\centering
\includegraphics[width=0.7\linewidth]{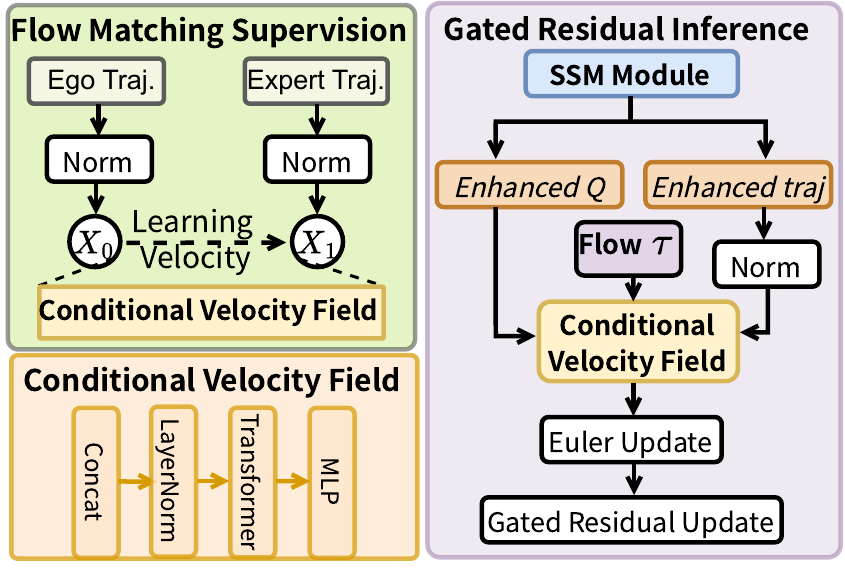}
\caption{\textbf{Flow Matching Supervision} learns a query-conditioned residual velocity field that transforms the SSM-enhanced planner trajectory toward the expert trajectory. During inference, \textbf{Gated Residual Refinement} integrates this field with a few Euler steps and uses a bounded residual gate to improve trajectory accuracy while preserving anchor-based planning stability.
} \label{fig:flow_refiner}
\end{figure}

\section{Related Work}
\subsection{End-to-End Autonomous Driving}
Existing studies have advanced E2E-AD through unified representations, efficient decoding, generative planning, VLA models, and world models. Representative methods include planning-oriented unified frameworks \cite{uniad,jiang2023vad,chen2024vadv2}, efficient planning decoders \cite{jia2023driveadapter,jia2025drivetransformer,sun2024sparsedrive,li2024hydra}, generative trajectory modeling \cite{liao2024diffusiondrive,xing2025goalflow,liu2025guideflow,diver}, reliable planning optimization \cite{shang2025drivedpo,ResAD}, large-model-based driving cognition \cite{autovla}, and world-model-based long-horizon reasoning \cite{song2026graphworld}. However, existing Temporal E2E-AD methods overlook temporal memory reliability. When scenes or commands change, reused history may conflict with the current intention and cause negative interference. Thus, long-horizon planning should selectively retain reliable memory while suppressing invalid history.

\subsection{State Space Models for Autonomous Driving}
State space models (SSMs) offer an efficient sequence modeling paradigm for capturing long-range dependencies through latent dynamical systems. Representative models such as  Mamba~\cite{gu2024mamba}, and Mamba-2~\cite{dao2024mamba2} improve long-sequence modeling via structured dynamics, selective state transitions, and hardware-aware linear-complexity computation. Beyond language modeling, SSMs have been extended to visual representation learning~\cite{zhu2024visionmamba,liu2024vmamba}, occupancy prediction~\cite{li2025occmamba}, and E2E-AD~\cite{yuan2024drama,su2026drivemamba,wang2025gmf,lu2025me}. Existing Mamba-based E2E-AD methods mainly focus on efficient temporal modeling, while the reliability of historical memory in long-horizon planning remains underexplored.

\section{Methodology}
\label{sec:methodology}


\subsection{Overview of MomADv2}
\label{sec:overallframework}
As shown in Figure~\ref{fig:framework}, MomADv2 inputs multi-view camera images and first extracts sparse scene representations for surrounding agents and map elements. An anchor-based motion and planning decoder then produces multi-modal motion predictions and initial ego planning candidates. Built upon the anchor-based planner, MomADv2 introduces two modules: a \textbf{Selective State-Space Memory Query Module} that filters and enhances reliable historical planning queries, and a \textbf{Flow-Matching Residual Refiner} that corrects planned trajectories with query-conditioned residual fields.

\subsection{Selective State-Space Planning Memory Query}

As shown in Figure~\ref{fig:framework}, the proposed \textbf{Selective State-Space Planning Memory Query Module (SSM-Q)} enhances the current planning query by selectively retrieving reliable historical planning states. Instead of directly propagating historical predictions, we maintain a token-keyed memory bank that stores raw planning queries and baseline command-specific candidate trajectories before temporal enhancement:
\begin{equation}
\mathcal{M}_t
=
\left\{
\chi_{t-k}
\mapsto
\left(
\mathbf{Q}^{\mathrm{raw}}_{t-k},
\mathbf{Y}^{\mathrm{base,cmd}}_{t-k},
c_{t-k}
\right)
\right\}_{k=1}^{K}.
\end{equation}
Here, $(c_{t-k}, \chi_{t-k})$ serves as the memory key, while
$(\mathbf{Q}^{\mathrm{raw}}_{t-k},
\mathbf{Y}^{\mathrm{base,cmd}}_{t-k})$ forms the associated value.
Specifically, $c_{t-k}$ denotes the high-level driving command, e.g.,
$\{\texttt{left}, \texttt{straight}, \texttt{right}\}$, and
$\chi_{t-k}$ is the temporal-continuity token.
For each command, the planner maintains $M_c$ candidate trajectories,
each paired with a corresponding raw planning query.
For example, when the current command is \texttt{left}, only the $M_c$
left-turn candidates and their paired queries from a temporally
continuous historical frame are eligible for retrieval, while candidates
associated with \texttt{straight} or \texttt{right} are excluded.
Storing pre-enhancement queries and baseline trajectories prevents
SSM-corrected outputs from being recursively written back into memory,
thereby avoiding self-amplified temporal drift.

\noindent\textbf{Reliable Historical Query Retrieval.}
Historical planning states are used only when they are command-consistent, temporally continuous, and available in memory:
\begin{equation}
m_{t,k}
=
\mathbb{I}
\left[
c_t=c_{t-k},
\;
\chi_t^{(-k)}=\chi_{t-k},
\;
\mathbf{Q}^{\mathrm{raw}}_{t-k}\neq \mathbf{0}
\right],
\end{equation}
where $m_{t,k}$ denotes the validity indicator for the historical state at time $t-k$. Since planning candidates are organized according to high-level driving commands, historical retrieval is restricted to the candidate subset associated with the current command. However, the ordering of candidate trajectories may vary across frames due to changes in scene context and planning ambiguity. Therefore, for each current command-specific planning candidate $j$, we identify the most trajectory-consistent historical candidate under the same command:
\begin{equation}
h^{*}_{t,k,j}
=
\arg\min_{h \in \{1,\dots,M_c\}}
D_{\mathrm{shift}}
\left(
\mathbf{Y}^{\mathrm{base,cmd}}_{t-k,h},
\mathbf{Y}^{\mathrm{base,cmd}}_{t,j}
\right).
\end{equation}
The corresponding historical query is retrieved as
\begin{equation}
\widehat{\mathbf{Q}}^{\mathrm{hist}}_{t,k,j}
=
\mathbf{Q}^{\mathrm{raw}}_{t-k}
\left[
\operatorname{idx}
\left(
c_t,
h^{*}_{t,k,j}
\right)
\right],
\end{equation}
where $\operatorname{idx}(c,h)=cM_c+h$ maps the command-specific candidate index to the global candidate index. Here, $D_{\mathrm{shift}}(\cdot,\cdot)$ denotes a temporally shifted trajectory distance that compensates for frame-to-frame trajectory offsets. This trajectory-level alignment enables each current planning candidate to retrieve the most relevant historical query, rather than relying on a fixed candidate index across time.

\noindent\textbf{Reliability-aware Temporal Modeling.}
The aligned historical queries are aggregated according to validity, trajectory similarity, and temporal decay:
\begin{equation}
\bar{\mathbf{Q}}^{\mathrm{hist}}_{t,j}
=
\sum_{k=1}^{K}
w_{t,k,j}
\widehat{\mathbf{Q}}^{\mathrm{hist}}_{t,k,j}.
\end{equation}
The aggregation weight is computed as
\begin{equation}
w_{t,k,j}
=
\frac{
\phi_{t,k,j}
}{
\sum_{k'=1}^{K}
\phi_{t,k',j}
+
\epsilon
},
\end{equation}
where
\begin{equation}
\phi_{t,k,j}
=
m_{t,k}
\exp
\left(
-\frac{d_{t,k,j}}{\alpha}
\right)
\exp
\left(
-\frac{k}{\beta}
\right).
\end{equation}
Here, $d_{t,k,j}$ is the aligned trajectory distance, while $\alpha$ and $\beta$ control the sensitivity to trajectory similarity and temporal decay, respectively.

We further compute a reliability gate to prevent unreliable historical states from perturbing the current planning query:
\begin{equation}
r_{t,j}
=
r^{\mathrm{cnt}}_t
\cdot
r^{\mathrm{dist}}_{t,j},
\end{equation}
where
\begin{equation}
r^{\mathrm{cnt}}_t
=
\min
\left(
\frac{
\sum_{k=1}^{K}m_{t,k}
}{
K_0
},
1
\right),
\end{equation}
and
\begin{equation}
r^{\mathrm{dist}}_{t,j}
=
\sigma
\left(
\frac{
d_0-\min_k d_{t,k,j}
}{
\gamma_d
}
\right).
\end{equation}
Here, $K_0$ denotes the minimum number of valid historical states required for reliable temporal modeling, $d_0$ is the distance threshold, and $\gamma_d$ controls the sharpness of the distance-based reliability gate. Thus, historical memory is activated only when sufficient valid and trajectory-consistent historical information is available.

For each command-specific planning candidate, the aligned historical queries and the current query are concatenated into a temporal sequence:
\begin{equation}
\mathbf{Z}^{0}_{t,j}
=
\operatorname{Concat}
\left(
\widehat{\mathbf{Q}}^{\mathrm{hist}}_{t,K,j},
\dots,
\widehat{\mathbf{Q}}^{\mathrm{hist}}_{t,1,j},
\mathbf{Q}_{t,j}
\right).
\end{equation}
The sequence is processed by a selective state-space query encoder composed of layer normalization, causal depthwise convolution, input-dependent selective scan, SiLU gating, and residual connection:
\begin{equation}
\mathbf{H}^{\mathrm{cur}}_{t,j}
=
\operatorname{SSMEncoder}
\left(
\mathbf{Z}^{0}_{t,j}
\right)[-1].
\end{equation}
The encoder models the temporal transition from historical planning intentions to the current planning intention while preserving candidate-wise planning semantics.

\noindent\textbf{Reliability-gated Residual Injection.}
The SSM-induced residual direction is computed as
\begin{equation}
\Delta \mathbf{Q}_{t,j}
=
\tanh
\left(
\mathbf{H}^{\mathrm{cur}}_{t,j}
-
\mathbf{Q}_{t,j}
\right).
\end{equation}
To avoid excessive perturbation to the original planner, we apply a reliability-gated and norm-constrained residual update:
\begin{equation}
\mathbf{U}_{t,j}
=
\operatorname{Clip}
\left(
g_{t,j}
\Delta \mathbf{Q}_{t,j},
\;
\rho
\left\|
\mathbf{Q}_{t,j}
\right\|_2
\right),
\end{equation}
where $g_{t,j}$ combines the valid-history indicator, the global residual gate, the reliability gate, the candidate-wise gate, and the residual scale. The coefficient $\rho$ limits the maximum magnitude of query perturbation. The enhanced planning query is obtained by
\begin{equation}
\widetilde{\mathbf{Q}}_{t,j}
=
\mathbf{Q}_{t,j}
+
\mathbf{U}_{t,j}.
\end{equation}

Finally, the enhanced planning query is fed into the planning refinement head:
\begin{equation}
\{
\boldsymbol{\pi}^{\mathrm{ssm}}_t,
\mathbf{Y}^{\mathrm{ssm}}_t,
\mathbf{s}^{\mathrm{ssm}}_t
\}
=
f_{\mathrm{ref}}
\left(
\widetilde{\mathbf{Q}}_t
\right),
\end{equation}
where $\boldsymbol{\pi}^{\mathrm{ssm}}_t$, $\mathbf{Y}^{\mathrm{ssm}}_t$, and $\mathbf{s}^{\mathrm{ssm}}_t$ denote the planning classification, trajectory regression, and planning status, respectively. 

\subsection{Flow-Matching Trajectory Residual Refiner}
\label{Sec:FlowRefinement}

We propose a \textbf{Flow-Matching Trajectory Residual Refiner (FM-Ref)} to improve ego trajectory accuracy and smoothness, as shown in Figure \ref{fig:flow_refiner}. Instead of generating trajectories from random noise as diffusion methods do, our refiner starts from planner-refined candidates and learns a continuous residual correction field toward the expert trajectory. This residual refinement preserves the stability of the anchor-based planner while enhancing local trajectory quality.

\noindent\textbf{Conditional Flow Field.}
Let
$
\mathbf{Y}^{\mathrm{ref}}_t
\in
\mathbb{R}^{B \times 1 \times N \times T \times 2}
$
denote the planner-refined candidate trajectories in displacement form, where $B$ is the batch size, $N$ is the number of planning candidate trajectories maintained by the planning head, and $T$ is the number of future timesteps. The trajectory $\mathbf{Y}^{\mathrm{ref}}_t$ is produced by the SSM-enhanced planning branch. Since the planning head predicts trajectory offsets, we first convert the displacement-form trajectories into the absolute trajectory space by cumulative summation:
\begin{equation}
\overline{\mathbf{Y}}^{\mathrm{ref}}_t
=
\operatorname{Cumsum}
\left(
\mathbf{Y}^{\mathrm{ref}}_t
\right),
\end{equation}
where $\operatorname{Cumsum}(\cdot)$ is performed along the temporal dimension. The normalized initial trajectory state is then defined as
\begin{equation}
\mathbf{x}_0
=
\frac{
\overline{\mathbf{Y}}^{\mathrm{ref}}_t
}{s},
\end{equation}
where $s$ is a trajectory scale factor. During the optimization of the flow branch, the planner-refined trajectory and the enhanced planning query are treated with stop-gradient to prevent the auxiliary flow objective from destabilizing the planner.

Given the enhanced planning query $\widetilde{\mathbf{Q}}_t$ as the conditional planning context, an intermediate trajectory state $\mathbf{x}_{\tau}$ as the variable to be refined, and a continuous flow time $\tau \in [0,1]$, we introduce a \textbf{Conditional Velocity Field} to predict a time-dependent residual velocity field. Specifically, the estimator first embeds the enhanced planning query, the trajectory state, and the flow time:
\begin{equation}
\mathbf{z}_{\tau}
=
\operatorname{Concat}
\left(
\phi_q
\left(
\widetilde{\mathbf{Q}}_t
\right),
\phi_x
\left(
\mathbf{x}_{\tau}
\right),
\phi_{\tau}
\left(
\tau
\right)
\right),
\end{equation}
where $\phi_q(\cdot)$, $\phi_x(\cdot)$, and $\phi_{\tau}(\cdot)$ denote the query projection, trajectory encoder, and time embedding function, respectively. The fused representation is processed by an MLP with LayerNorm and a Transformer encoder:
\begin{equation}
\mathbf{h}_{\tau}
=
\operatorname{TransEnc}
\left(
\operatorname{LN}
\left(
\operatorname{MLP}
\left(
\mathbf{z}_{\tau}
\right)
\right)
\right).
\end{equation}
Finally, an MLP-based velocity head predicts the residual velocity field:
\begin{equation}
\mathbf{v}_{\theta}
=
f_{\theta}
\left(
\widetilde{\mathbf{Q}}_t,
\mathbf{x}_{\tau},
\tau
\right)
=
\operatorname{MLP}_{\mathrm{vel}}
\left(
\mathbf{h}_{\tau}
\right).
\end{equation}
This estimator conditions the residual correction field on both the planning intention encoded by $\widetilde{\mathbf{Q}}_t$ and the geometric state of the current trajectory.

\noindent\textbf{Flow Matching Supervision.}
During training, the expert ego trajectory
$
\mathbf{Y}^{\star}_t
$
is also converted into the absolute trajectory space and normalized:
\begin{equation}
\overline{\mathbf{Y}}^{\star}_t
=
\operatorname{Cumsum}
\left(
\mathbf{Y}^{\star}_t
\right),
\end{equation}
\begin{equation}
\mathbf{x}_1
=
\frac{
\overline{\mathbf{Y}}^{\star}_t
}{s}.
\end{equation}
Here, $\mathbf{Y}^{\star}_t$ denotes the expert trajectory in displacement form. In supervised training, the loss is computed on the planning candidate selected by the planning sampler, while the candidate index is omitted for notational simplicity.

We sample $\tau \sim \mathcal{U}(0,1)$ and construct an intermediate trajectory state along the linear path from the planner-refined trajectory to the expert trajectory:
\begin{equation}
\mathbf{x}_{\tau}
=
(1-\tau)
\mathbf{x}_0
+
\tau
\mathbf{x}_1.
\end{equation}
The target velocity is defined as the residual displacement between the two endpoints in the normalized absolute trajectory space:
\begin{equation}
\mathbf{v}^{\star}
=
\mathbf{x}_1
-
\mathbf{x}_0.
\end{equation}
The flow-matching objective supervises the predicted residual velocity field:
\begin{equation}
\mathcal{L}_{\mathrm{FM}}
=
\left\|
f_{\theta}
\left(
\widetilde{\mathbf{Q}}_t,
\mathbf{x}_{\tau},
\tau
\right)
-
\mathbf{v}^{\star}
\right\|_2^2.
\end{equation}
Through this objective, the velocity estimator learns a query-conditioned residual transport field that moves the planner-refined trajectory toward the expert trajectory in the normalized absolute trajectory space.

\noindent\textbf{Gated Residual Inference.}
During inference, no expert trajectory is available. Therefore, we start from the normalized planner-refined trajectory $\mathbf{x}_0$ and progressively update it by Euler integration:
\begin{equation}
\mathbf{x}_{k+1}
=
\mathbf{x}_{k}
+
\frac{1}{S}
f_{\theta}
\left(
\widetilde{\mathbf{Q}}_t,
\mathbf{x}_{k},
\tau_k
\right),
\end{equation}
where $\tau_k=(k+0.5)/S$, and $S$ denotes the number of integration steps. In our implementation, we set $S=2$ for efficient inference.

To preserve the stability of the anchor-based planner and avoid overly aggressive corrections, we introduce a learnable bounded residual gate:
\begin{equation}
\eta
=
\eta_{\min}
+
\eta_{\max}
\sigma
\left(
g_{\mathrm{flow}}
\right),
\end{equation}
where $\eta_{\min}$ provides a small base correction ratio and $\eta_{\max}$ bounds the maximum residual strength. The final flow-refined trajectory in the absolute trajectory space is obtained by applying the gated residual correction to the initial planner-refined trajectory:
\begin{equation}
\overline{\mathbf{Y}}^{\mathrm{flow}}_t
=
s
\left[
\mathbf{x}_0
+
\eta
\left(
\mathbf{x}_S
-
\mathbf{x}_0
\right)
\right].
\end{equation}
When required by the planning decoder, the absolute-form trajectory is converted back into the displacement form:
\begin{equation}
\mathbf{Y}^{\mathrm{flow}}_t
=
\operatorname{Diff}
\left(
\overline{\mathbf{Y}}^{\mathrm{flow}}_t
\right),
\end{equation}
where $\operatorname{Diff}(\cdot)$ denotes the inverse operation of cumulative summation along the temporal dimension. Therefore, the proposed module performs gated residual refinement rather than directly replacing the planner output, which maintains the stability of anchor-based planning while improving local trajectory accuracy and smoothness.

\noindent\textbf{Training Objectives.}
During training, the planner branch is supervised by the standard planning objective and the temporal refinement losses introduced by the SSM-enhanced query module. In addition to the flow-matching objective, the flow-refined trajectory is further optimized with a regression loss:
\begin{equation}
\mathcal{L}_{\mathrm{flow}}
=
\ell_{\mathrm{reg}}
\left(
\mathbf{Y}^{\mathrm{flow}}_t,
\mathbf{Y}^{\star}_t
\right).
\end{equation}
In practice, the flow output can be supervised in both displacement and cumulative trajectory spaces:
\begin{equation}
\mathcal{L}_{\mathrm{flow}}
=
\lambda_{\Delta}
\ell_{\mathrm{reg}}
\left(
\mathbf{Y}^{\mathrm{flow}}_t,
\mathbf{Y}^{\star}_t
\right)
+
\lambda_{\mathrm{cum}}
\ell_{\mathrm{reg}}
\left(
\overline{\mathbf{Y}}^{\mathrm{flow}}_t,
\overline{\mathbf{Y}}^{\star}_t
\right),
\end{equation}
where $\lambda_{\Delta}$ and $\lambda_{\mathrm{cum}}$ balance displacement-form and cumulative-form supervision.

The overall objective for the planning refinement stage is formulated as
\begin{equation}
\mathcal{L}
=
\mathcal{L}_{\mathrm{plan}}
+
\mathcal{L}_{\mathrm{SSM}}
+
\lambda_{\mathrm{flow}}
\mathcal{L}_{\mathrm{flow}}
+
\lambda_{\mathrm{FM}}
\mathcal{L}_{\mathrm{FM}},
\end{equation}
where $\mathcal{L}_{\mathrm{plan}}$ denotes the standard planning loss, $\mathcal{L}_{\mathrm{SSM}}$ denotes the temporal trajectory refinement losses associated with the SSM-enhanced planning branch, and $\lambda_{\mathrm{flow}}$ and $\lambda_{\mathrm{FM}}$ are balancing weights for the flow-refined regression loss and the flow-matching objective, respectively. During inference, $\mathbf{Y}^{\mathrm{flow}}_t$ is used as the final ego planning trajectory.

\begin{table*}[t]
\scriptsize
\centering
\caption[]{Planning results for \textbf{6-second long-horizon planning} on the \textbf{nuScenes} validation set.}
\renewcommand\arraystretch{0.7}
\setlength{\tabcolsep}{1.99mm}
  \resizebox{\linewidth}{!}{
\begin{tabular}{l c ccccccc  ccccccc }
\toprule
\multirow{2}{*}{$\operatorname{Method}$} &\multirow{2}{*}{$\operatorname{Venue}$} & \multicolumn{7}{c}{$\operatorname{L2\ (m)}\downarrow$} & \multicolumn{7}{c}{$\operatorname{Col.\ Rate\ (\%)}\downarrow$}\\
\cmidrule(lr){3-9} \cmidrule(lr){10-16}
&& 1s & 2s & 3s & 4s & 5s & 6s& $\operatorname{Avg.}$ & 1s & 2s & 3s & 4s & 5s & 6s &$\operatorname{Avg.}$ \\
\midrule
$\operatorname{UniAD}$~\cite{uniad}&CVPR'23 &0.47& 0.91& 1.35 &1.91& 2.47 &3.07&\cellcolor{red!5}1.70& 0.25& 0.36& 0.61 & 0.99 &1.64& 2.51&\cellcolor{red!5}1.06 \\
$\operatorname{SparseDrive}$~\cite{sun2024sparsedrive}&CVPR'23 &0.43& 0.87& 1.23 &1.75& 2.32 &2.95&\cellcolor{red!5}1.59& 0.19& 0.31& 0.56 & 0.87 &1.54& 2.33&\cellcolor{red!5}0.97 \\
$\operatorname{MomAD}$~\cite{momad}&CVPR'25&0.41& 0.85& 1.13 &1.67& 1.98& 2.45&\cellcolor{red!5}1.42&0.17& 0.30& 0.54 & 0.83& 1.43 &2.13&\cellcolor{red!5}0.90\\
$\operatorname{LAW}$~\cite{li2024enhancing_law}&ICLR'25&0.40& 0.87& 1.16 &1.71& 2.03& 2.61&\cellcolor{red!5}1.46&0.19& 0.33& 0.57 & 0.86& 1.51 &2.31&\cellcolor{red!5}0.96\\
$\operatorname{Epona}$~\cite{zhang2025epona}&ICCV'25&0.39& 0.91& 1.17 &1.73& 2.02& 2.75&\cellcolor{red!5}1.50&0.14& 0.18& 0.45 & 0.74& 1.48 &2.23&\cellcolor{red!5}0.87\\
$\operatorname{DIVER}$~\cite{diver}&TPAMI'26&0.38& 0.75& 1.10& 1.53& 1.98& 2.49& \cellcolor{red!5}1.37& 0.13& 0.31& 0.44& 0.80& 1.41& 2.11& \cellcolor{red!5}0.87\\
$\operatorname{GuideFlow}$~\cite{liu2025guideflow}&CVPR'26&0.42& 0.83& 1.21& 1.73& 2.05& 2.63& \cellcolor{red!5}1.48& 0.12& 0.22 &0.42& 0.79& 1.44& 2.15 &\cellcolor{red!5}0.86\\
\cellcolor{red!5}$\operatorname{MomADv2 (Ours)}$&\cellcolor{red!5}-&\cellcolor{red!5}\textbf{0.28}& \cellcolor{red!5}\textbf{0.54} & \cellcolor{red!5}\textbf{0.89} &\cellcolor{red!5}\textbf{1.32} &\cellcolor{red!5}\textbf{1.82} & \cellcolor{red!5}\textbf{2.40}&\cellcolor{red!5}\textbf{1.21}&\cellcolor{red!5}\textbf{0.00}& \cellcolor{red!5}\textbf{0.12}&\cellcolor{red!5}\textbf{0.33}& \cellcolor{red!5}\textbf{0.72}& \cellcolor{red!5}\textbf{1.33} &\cellcolor{red!5}\textbf{2.03}&\cellcolor{red!5}\textbf{0.76}\\
\bottomrule
\end{tabular}}
\label{tab_nuscenes_planning_6s}
\end{table*}

\begin{table}[t]
\scriptsize
\centering
\addtolength{\tabcolsep}{0.1pt}
\caption{$\operatorname{\textbf{Open-Loop}}$ and $\operatorname{\textbf{Closed-Loop}}$ results on $\operatorname{\textbf{Bench2Drive}}$ (V0.0.3) under base training set.
* denotes expert feature distillation. 
`DS' denotes Driving Score, `SR' denotes Success Rate, `Effi' denotes Efficiency, and `Comf' denotes Comfortness.
}
\renewcommand\arraystretch{0.7}
\tabcolsep=0.5mm
\resizebox{\linewidth}{!}{
\begin{tabular}{l l c cccc}
\toprule
\multirow{2}{*}{$\operatorname{Method}$} 
& \multirow{2}{*}{$\operatorname{Venue}$} 
& \multicolumn{1}{c}{$\operatorname{Open-loop}$} 
& \multicolumn{4}{c}{$\operatorname{Closed-loop}$} \\
\cmidrule(lr){3-3}\cmidrule(lr){4-7}
&& $\operatorname{Avg. L2}\downarrow$ 
& $\operatorname{DS}\uparrow$
& $\operatorname{SR\ (\%)}\uparrow$ 
& $\operatorname{Effi}\uparrow$  
& $\operatorname{Comf}\uparrow$ \\
\midrule
DriveAdapter* \cite{jia2023driveadapter} & ICCV'23 & 1.01 & 64.22 & 33.08 & 70.22 & 16.01 \\
DriveDPO \cite{shang2025drivedpo} & NeurIPS'25 & - & 62.02 & 30.62 & - & - \\
Raw2Drive \cite{Raw2Drive} & NeurIPS'25 & - & 71.36 & \textbf{50.24} & - & - \\
DriveTrans* \cite{jia2025drivetransformer} & ICLR'25 & \textbf{0.62} & 63.46 & 35.01 & \textbf{100.64} & 20.78 \\
WoTE* \cite{wote} & ICCV'25 & - & 61.71 & 31.36 & - & - \\
DIVER*  \cite{diver} & TPAMI'26 & 1.11 & 68.90 & 36.75 & 72.34 & 22.34 \\

\cellcolor{red!5}${\operatorname{MomADv2}}$ (Ours) * 
& \cellcolor{red!5} 
& \cellcolor{red!5}0.77 
& \cellcolor{red!5}\textbf{78.82}
& \cellcolor{red!5}46.50
& \cellcolor{red!5}83.41
& \cellcolor{red!5}\textbf{31.23} \\
\midrule
\midrule
UniAD-Base \cite{uniad} & CVPR'23 & \textbf{0.73} & 45.81 & 16.36 & 129.21 & 43.58 \\
$\operatorname{VAD}$ \cite{jiang2023vad} & ICCV'23 & 0.91 & 42.35 & 15.00 & 157.94 & 46.01 \\
GenAD \cite{zheng2024genad} & ECCV'24 & - & 44.81 & 15.90 & - & - \\
$\operatorname{MomAD}$ \cite{momad} & CVPR'25 & 0.82 & 47.91 & 18.11 & \textbf{174.91} & 51.20 \\
$\operatorname{DIVER}$ \cite{diver} & TPAMI'26 & 1.13 & 47.95 & 19.47 & 164.66 & 51.28 \\
\cellcolor{red!5}${\operatorname{MomADv2}}$ (Ours)
& \cellcolor{red!5}
& \cellcolor{red!5}0.76 
& \cellcolor{red!5}\textbf{52.32}
& \cellcolor{red!5}\textbf{24.24}
& \cellcolor{red!5}169.09
& \cellcolor{red!5}\textbf{55.11} \\
\bottomrule
\end{tabular}
}
\label{tab_b2d}
\end{table}

\begin{table}[t]
\centering
  \caption{\textcolor{black}{Comparison on planning-oriented \textbf{NAVSIMv1} navtest split with $\operatorname{\textbf{Closed-Loop}}$ metrics.}}
\renewcommand\arraystretch{0.7}
  \tabcolsep=0.6mm 
  \resizebox{\linewidth}{!}{
    \begin{tabular}{l c c c c c c c}
    \toprule
    \multicolumn{1}{l}{Method}&Venue&  NC$\uparrow$& DAC$\uparrow$& TTC$\uparrow$& Comf.$\uparrow$& EP$\uparrow$& PDMS$\uparrow$ \\
\midrule
TransFuser~\cite{TransFuser}&TPAMI'22 &  97.7 & 92.8 & 92.8 & 100 & 79.2 & 84.0 \\
VADv2~\cite{chen2024vadv2}&ICLR'26 &  97.2 & 89.1 & 91.6 & 100 & 76.0 & 80.9 \\
GoalFlow~\cite{xing2025goalflow}&CVPR’25   & 98.3 & 93.8 & 94.3 & 100 & 79.8 & 85.7 \\
Hydra-MDP~\cite{li2024hydra}&Arxiv’24 &  98.3 & 96.0 & 94.6 & 100 & 78.7 & 86.5 \\
FUMP~\cite{liu2025fump}&Arxiv’25 &  98.1 & 96.2 & 94.2 & 100 & 82.0 & 87.8 \\
DiffusionDrive~\cite{liao2024diffusiondrive}&CVPR’25 & 98.2 & 96.2 & 94.7 & 100 & 82.2 & 88.1 \\
DIVER~\cite{diver}&TPAMI’26 &  98.5 & 96.5 & 94.9 & 100 & 82.6 & 88.3 \\
DriveSuprim~\cite{yao2025drivesuprim} &AAAI’26&  97.8 & \textbf{97.3} & 93.6 & 100 & \textbf{86.7} & \textbf{89.9} \\
\cellcolor{red!5}$\operatorname{MomADv2\ (Ours)}$ 
& \cellcolor{red!5}-
& \cellcolor{red!5}\textbf{98.7}
& \cellcolor{red!5}96.8
& \cellcolor{red!5}\textbf{95.4}
& \cellcolor{red!5}100
& \cellcolor{red!5}84.5
& \cellcolor{red!5}\textbf{89.9}\\
            \bottomrule
    \end{tabular}}
\label{tab_NAVSIMv1}
\end{table}

\begin{table}[t]
\centering
\caption{Comparison with SOTA methods on the \textbf{NAVSIMv2 navhard} split~\cite{navsimv2}.}
\renewcommand\arraystretch{0.9}
\tabcolsep=0.3mm
\resizebox{\linewidth}{!}{
\begin{tabular}{l c ccccc ccccc}
\toprule
\multicolumn{1}{l}{Method} & Stage 
& NC$\uparrow$ & DAC$\uparrow$ & DDC$\uparrow$ & TL$\uparrow$ & EP$\uparrow$ 
& TTC$\uparrow$ & LK$\uparrow$ & HC$\uparrow$ & EC$\uparrow$ & EPDMS$\uparrow$ \\
\midrule
\multirow{2}{*}{TransFuser} 
&Stage1 & 96.2 & 79.5 & 99.1 & 99.5 & 84.1 & 95.1 & 94.2 & 97.5 & 79.1 & \multirow{2}{*}{23.1} \\
&Stage2 & 77.7 & 70.2 & 84.2 & 98.0 & 85.1 & 75.6 & 45.4 & 95.7 & 75.9 &  \\
\arrayrulecolor{gray!25}\cmidrule[0.01em]{1-12}\arrayrulecolor{black}
\multirow{2}{*}{DiffusionDrive} 
&Stage1 & 96.0 & 79.7 & 97.4 & 99.5 & 81.3 & 93.1 & 90.8 & 96.8 & 73.8 & \multirow{2}{*}{24.2} \\
&Stage2 & 82.1 & 72.2 & 88.5 & 98.7 & 85.1 & 78.8 & 49.2 & 89.3 & 71.2 &  \\
\arrayrulecolor{gray!25}\cmidrule[0.01em]{1-12}\arrayrulecolor{black}
\multirow{2}{*}{GuideFlow} 
&Stage1 & 96.6 & 80.5 & 96.3 & 99.3 & 82.3 & 94.9 & 91.5 & 97.7 & 67.8 & \multirow{2}{*}{27.1} \\
&Stage2 & 87.3 & 76.7 & 88.8 & 99.2 & 84.3 & 85.1 & 49.7 & 93.1 & 44.5 &  \\
\arrayrulecolor{gray!25}\cmidrule[0.01em]{1-12}\arrayrulecolor{black}
\multirow{2}{*}{\textbf{MomADv2 (Ours)}} 
&Stage1 & 97.7 & 92.7 & 97.6 & 98.7 & 73.4 & 98.0 & 92.5 & 96.1 & 51.0 & \multirow{2}{*}{\textbf{39.5}} \\
&Stage2 & 85.9 & 87.0 & 87.1 & 97.5 & 77.2 & 83.8 & 50.1 & 95.4 & 53.3 &  \\
\bottomrule
\end{tabular}}
\label{tab_navsimv2_navhard}
\end{table}

\begin{table}[t]
\centering
\caption{Comparison with SOTA methods on the \textbf{NAVSIMv2 navtest} split~\cite{navsimv2}.}
\renewcommand\arraystretch{0.9}
\tabcolsep=0.3mm
\resizebox{\linewidth}{!}{
\begin{tabular}{l c ccccc cccc}
\toprule
\multicolumn{1}{l}{Method} & NC$\uparrow$ & DAC$\uparrow$ & DDC$\uparrow$ & TL$\uparrow$ & EP$\uparrow$ & TTC$\uparrow$ & LK$\uparrow$ & HC$\uparrow$ & EC$\uparrow$ & EPDMS$\uparrow$ \\
\midrule
TransFuser~\cite{TransFuser} & 96.9 & 89.9 & 97.8 & 99.7 & 87.1 & 95.4 & 92.7 & 98.3 & 87.2 & 76.7 \\
DiffusionDrive~\cite{liao2024diffusiondrive} & 98.2 & 95.9 & 99.4 & \textbf{99.8} & 87.5 & 97.3 & 96.8 & 98.3 & 87.7 & 84.5 \\
Hydra-MDP++~\cite{li2025hydraplus} & 97.2 & 97.5 & 99.4 & 99.6 & 83.1 & 96.5 & 94.4 & 98.2 & 70.9 & 81.4 \\
DriveSuprim~\cite{yao2025drivesuprim} & 97.5 & 96.5 & 99.4 & 99.6 & 88.4 & 96.6 & 95.5 & 98.3 & 77.0 & 83.1 \\
DiffusionDriveV2~\cite{zou2025diffusiondrivev2} & 97.7 & 96.6 & 99.2 & \textbf{99.8} & 88.9 & 97.2 & 96.0 & 97.8 & 91.0 & 87.5 \\
DriveWorld-VLA~\cite{liu2026driveworld} & \textbf{98.6} & \textbf{99.1} & 99.6 & \textbf{99.8} & 87.4 & 97.9 & \textbf{97.0} & 97.8 & 78.6 & 86.8 \\
DriveVLA-W0~\cite{li2025drivevla} & 98.5 & \textbf{99.1} & 98.0 & 99.7 & 86.4 & \textbf{98.1} & 93.2 & 97.9 & 58.9 & 86.1 \\
Recogdrive~\cite{li2025recogdrive} & 98.3 & 95.2 & 98.3 & \textbf{99.8} & 87.1 & 97.5 & 96.6 & \textbf{99.5} & 86.5 & 83.6 \\
\rowcolor{red!5} MomADv2 (Ours) & 98.3 & 97.4 & \textbf{99.7} & 99.3 & \textbf{89.6} & 97.8 & 95.5 & 98.6 & \textbf{91.7} & \textbf{87.9} \\
\bottomrule
\end{tabular}}
\label{tab_navsimv2_navtest}
\end{table}

\section{Experiments}

\subsection{Datasets and Metrics}
\noindent \textbf{Open-Loop.} Our MomADv2 is evaluated on 
nuScenes dataset. The \textbf{nuScenes}~\cite{nuscenes} (NuS) dataset consists of 1,000 driving scenes, each providing six synchronized camera images and LiDAR point clouds with a 360$^\circ$ field of view. In our experiments, we only use multi-view image data as model inputs. 
For nuScenes, we report the standard \(L_2\) displacement error and Collision Rate to jointly evaluate long-horizon planning accuracy and safety.

\noindent \textbf{Closed-Loop.} We evaluate MomADv2 on Bench2Drive and NAVSIM. \textbf{Bench2Drive}~\cite{jia2024bench2drive} is a closed-loop evaluation benchmark built upon CARLA Leaderboard 2.0 for E2E-AD. It provides an official training set, from which we use the base split (1,000 clips) to ensure fair comparison with prior methods, and an evaluation set consisting of 220 predefined routes. Following the official protocol, we report Driving Score (DS) and Success Rate (SR).
\textbf{NAVSIMv1/2}~\cite{navsimv1,navsimv2} is a planning benchmark derived from OpenScene, featuring multi-view camera and LiDAR data for 360$^\circ$ perception, with annotations at 2~Hz including HD maps and object bounding boxes. It adopts non-reactive simulation with closed-loop evaluation for comprehensive planning assessment.

\subsection{Implementation Details} 
We compare \textbf{MomADv2} with representative end-to-end autonomous driving baselines, including MomAD~\cite{momad} and TransFuser~\cite{TransFuser}. MomAD is used as the baseline on Bench2Drive, nuScenes, and Adv-nuSc, while TransFuser is adopted on NAVSIM following the official Navtrain split. For nuScenes and Adv-nuSc, we use a ResNet-50~\cite{resnet} backbone with an input resolution of $256 \times 704$, a 55 m detection range, a $60 \times 30$ m mapping range, and 6 motion trajectory modes. For Bench2Drive, we use a ResNet-50 backbone with 6 decoder layers, $640 \times 352$ input resolution, 900 agent queries, 100 map queries, and 480 planning queries. For NAVSIM, we follow TransFuser with the same perception modules and ResNet-34 backbone for fair comparison.
All experiments are conducted on 8 NVIDIA RTX 4090 GPUs. The batch size is 48 for nuScenes and Bench2Drive and 64 for NAVSIM. Models are trained for 20 epochs on nuScenes, 2 epochs on Bench2Drive, and 100 epochs on NAVSIM, taking approximately 10.3, 49.1, and 3.0 hours, respectively. 

\subsection{Main Results}

\noindent\textbf{Long-Horizon Planning on nuScenes (Open-Loop).}
Table~\ref{tab_nuscenes_planning_6s} reports open-loop 6-second planning results on the nuScenes validation set.
Existing E2E-AD methods suffer from increasing errors and collision rates as the horizon extends.
In contrast, \textbf{MomADv2} achieves the best performance across all timestamps, with the lowest average $L_2$ error of 1.21 m and collision rate of 0.76\%.
Compared with the strongest prior method, it improves average $L_2$ error by 11.7\% and collision rate by 11.6\%, demonstrating the effectiveness of selective state-space memory and flow-matching residual refinement.

\noindent\textbf{Bench2Drive (Closed-Loop).}
Table~\ref{tab_b2d} reports Bench2Drive results. \textbf{MomADv2} achieves strong closed-loop performance, reaching 78.82 DS and the best Comfortness of 31.23. It also improves MomAD from 47.91 to 52.32 DS and from 18.11\% to 24.24\% SR, demonstrating more reliable driving via effective temporal memory modeling.

\noindent\textbf{NAVSIMv1 navtest (Closed-Loop).}
Table~\ref{tab_NAVSIMv1} reports closed-loop results on NAVSIMv1 navtest. \textbf{MomADv2} achieves the highest NC of 98.7 and TTC of 95.4 among the compared methods, with 96.8 DAC, 100 Comfort, and 84.5 EP. Its PDMS of 89.9 ties DriveSuprim for the best score and exceeds TransFuser by 5.9 points.

\noindent\textbf{NAVSIMv2 navhard (Closed-Loop).}
Table~\ref{tab_navsimv2_navhard} reports results on the challenging NAVSIMv2 navhard split. \textbf{MomADv2} achieves the best overall EPDMS of 39.5, clearly outperforming TransFuser, DiffusionDrive, and GuideFlow. It maintains strong Stage-2 performance, especially on DAC and TTC, showing better robustness under long-horizon closed-loop interactions.

\noindent\textbf{NAVSIMv2 navtest (Closed-Loop).}
Table~\ref{tab_navsimv2_navtest} reports the results on the NAVSIMv2 navtest split.
\textbf{MomADv2} achieves the best EPDMS of 87.9 and ranks first on DDC, EP, and EC.
These results demonstrate that reliable temporal memory improves long-horizon planning consistency and closed-loop decision quality.

\begin{table}[t]
\centering
\caption{\textcolor{black}{Ablation study of the two core modules on the \textbf{NAVSIMv1} navtest split. 
SSM-Q denotes the Selective State-Space Planning Memory Query Module, and FM-Ref denotes the Flow-Matching Trajectory Residual Refiner.}}
\renewcommand\arraystretch{0.75}
\tabcolsep=2.6mm
\resizebox{\linewidth}{!}{
\begin{tabular}{c c c c c c c c}
\toprule
 SSM-Q & FM-Ref & NC$\uparrow$ & DAC$\uparrow$ & TTC$\uparrow$ & Comf.$\uparrow$ & EP$\uparrow$ & PDMS$\uparrow$ \\
\midrule
 & 
& 97.7 & 92.8 & 92.8 & 100 & 79.2 & 84.0 \\

\checkmark & 
& 98.2 & 94.5 & 94.0 & 100 & 82.1 & 86.9 \\

\rowcolor{red!5}
 \checkmark & \checkmark 
& \textbf{98.7} & \textbf{96.8} & \textbf{95.4} & 100 & \textbf{84.5} & \textbf{89.9} \\
\bottomrule
\end{tabular}}
\label{tab_ablation_modules_navsimv1}
\end{table}

\begin{table}[t]
\scriptsize
\centering
\caption{
Ablation study of reliable temporal memory on the \textbf{nuScenes} validation set.
All MomADv2 variants keep the FM-Ref and only differ in the historical memory strategy.
}
\renewcommand\arraystretch{0.75}
\setlength{\tabcolsep}{0.2mm}
\resizebox{\linewidth}{!}{
\begin{tabular}{l c ccccccc}
\toprule
\multirow{2}{*}{Method} 
& \multirow{2}{*}{Memory Strategy}
& \multicolumn{7}{c}{Col. Rate (\%)$\downarrow$} \\
\cmidrule(lr){3-9}
& & 1s & 2s & 3s & 4s & 5s & 6s & Avg. \\
\midrule
MomAD
& Historical query reuse
& 0.17 & 0.30 & 0.54 & 0.83 & 1.43 & 2.13 & 0.90 \\

\midrule
MomADv2
& No historical memory
& 0.16 & 0.29 & 0.52 & 0.86 & 1.48 & 2.18 & 0.92 \\

MomADv2
& Naive historical query fusion
& 0.18 & 0.34 & 0.60 & 0.94 & 1.62 & 2.35 & 1.01 \\

MomADv2
& Command-aware memory
& 0.12 & 0.25 & 0.47 & 0.82 & 1.46 & 2.19 & 0.89 \\

MomADv2
& + Temporal-continuity filtering
& 0.08 & 0.21 & 0.41 & 0.78 & 1.39 & 2.12 & 0.83 \\

MomADv2
& + Trajectory-level alignment
& 0.04 & 0.16 & 0.37 & 0.75 & 1.35 & 2.07 & 0.79 \\

\rowcolor{red!5}
MomADv2
& Full selective memory
& \textbf{0.00} & \textbf{0.12} & \textbf{0.33} & \textbf{0.72} & \textbf{1.33} & \textbf{2.03} & \textbf{0.76} \\
\bottomrule
\end{tabular}}
\label{tab_ablation_reliable_memory}
\end{table}

\begin{table}[t]
\centering
\caption{\textcolor{black}{Ablation study of historical memory length $K$ on the \textbf{NAVSIMv1} navtest split. 
$K$ denotes the number of historical planning states used in the SSM-Q.}}
\renewcommand\arraystretch{0.75}
\tabcolsep=4.6mm
\resizebox{\linewidth}{!}{
\begin{tabular}{c c c c c c c}
\toprule
$K$ & NC$\uparrow$ & DAC$\uparrow$ & TTC$\uparrow$ & Comf.$\uparrow$ & EP$\uparrow$ & PDMS$\uparrow$ \\
\midrule
1 
& 98.4 & 95.4 & 94.7 & 100 & 82.4 & 88.2 \\

2 
& \textbf{98.7} & 96.3 & 95.0 & 100 & 82.8 & 89.1 \\
\rowcolor{red!5}
4 
& \textbf{98.7} & \textbf{96.8} & \textbf{95.4} & 100 & \textbf{84.5} & \textbf{89.9} \\
6 
& 98.5 & 95.7 & 94.6 & 100 & 82.3 & 88.5 \\

8 
& 98.1 & 95.0 & 94.1 & 100 & 81.9 & 87.7 \\
\bottomrule
\end{tabular}}
\label{tab_ablation_history_length}
\end{table}

\begin{figure}[t]
\includegraphics[width=1\linewidth]{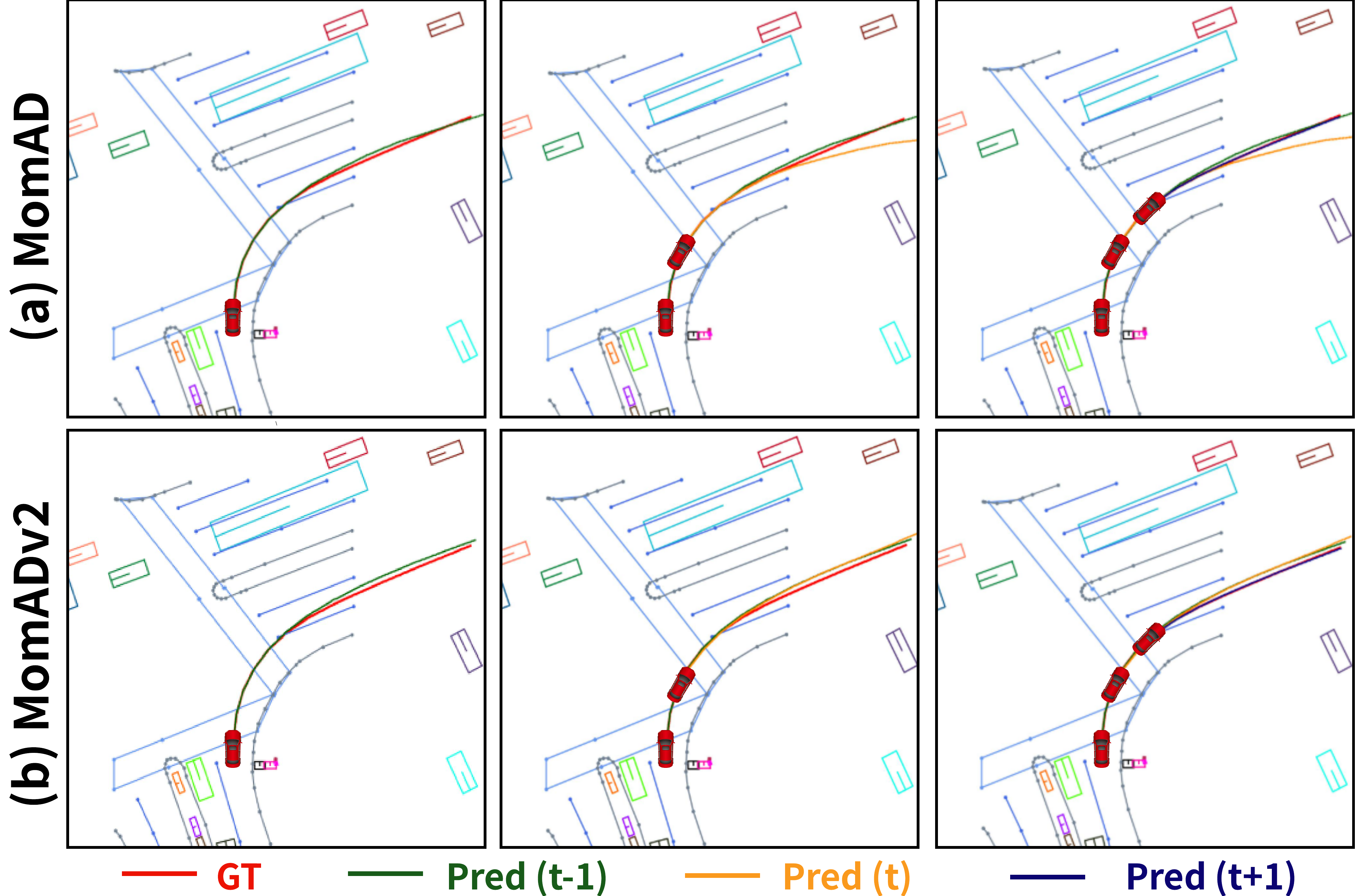}
\caption[ ]{\textbf{Visualization of MomADv2 on nuScenes dataset.}}
\label{fig:vis_planning_nus}
\end{figure}

\subsection{Ablation Studies}
\noindent\textbf{Roles of Different Modules in MomADv2.}
Table~\ref{tab_ablation_modules_navsimv1} validates the effectiveness of SSM-Q and FM-Ref.
SSM-Q improves PDMS from 84.0 to 86.9, showing the benefit of reliable temporal memory.
With FM-Ref, PDMS further increases to 89.9, demonstrating that residual trajectory refinement provides complementary gains.

\noindent\textbf{Ablation on Reliable Temporal Memory.}
Table~\ref{tab_ablation_reliable_memory} compares historical memory strategies.
The full selective strategy achieves the lowest collision rate of 0.76\%, outperforming naive fusion and partial filtering variants.
This confirms that reliable memory selection is essential for safe long-horizon planning.

\noindent\textbf{Ablation on Historical Memory Length.}
Table~\ref{tab_ablation_history_length} studies the effect of memory length $K$.
$K=4$ achieves the best PDMS of 89.9, while longer histories gradually degrade performance.
This suggests that moderate and reliable memory is more beneficial than excessively long historical context.

\subsection{Visualization}
Figure~\ref{fig:vis_planning_nus} visualizes consecutive planning results of MomAD and MambaFlow. 
Compared with MomAD, MomADv2 generates more temporally consistent trajectories across $t\!-\!1$, $t$, and $t\!+\!1$, with predictions closer to the ground truth. 
 Additional visualizations are provided in the appendix.

\section{Conclusion}
We propose \textbf{MomADv2}, a reliable state-space memory framework that selectively preserves useful history while suppressing stale or command-inconsistent memory.
Specifically, we introduce the Selective State-Space Planning Memory Query Module, which filters historical planning queries according to temporal continuity and command consistency, selects command-relevant planning modes, and models the evolution of planning intentions through a selective state-space mechanism. We further propose the Flow-Matching Trajectory Residual Refiner, which learns a continuous residual correction field to mitigate local trajectory deviations and accumulated planning errors while preserving the stability of anchor-based planning. Extensive experiments on NAVSIM, Bench2Drive, and nuScenes demonstrate that MomADv2 consistently improves long-horizon planning accuracy, temporal consistency, and driving safety.

\noindent\textbf{Limitation and Future Work.}
MomADv2 relies on predefined driving commands for memory selection and does not explicitly model uncertainty in future scene evolution. Future work will investigate command-free intention inference, uncertainty-aware memory selection, and world-model-based future reasoning.

\section{Acknowledgments}
This work was supported by the National Natural Science Foundation of China under Grant No. 52502496, the Beijing Natural Science Foundation under Grant No. L2609087, and the Natural Science Foundation of Chongqing, China under Grant No. CSTB2025NSCQ-GPX0413.

\bibliography{aaai2027}

\setcounter{secnumdepth}{2}
\appendix

\section{Appendix}

\subsection{Broader Impacts}
MomADv2 has the potential to improve the safety and reliability of autonomous driving by enhancing long-horizon planning consistency and suppressing unreliable temporal memory. Its selective state-space memory mechanism preserves useful historical context, while the flow-matching trajectory refiner improves trajectory accuracy and smoothness. These capabilities may support safer interactions, more stable decision-making, and greater robustness in complex traffic environments. More broadly, the proposed reliable temporal modeling paradigm may also benefit other sequential decision-making systems in robotics and embodied intelligence.

\subsection{Contributions}
Our contributions are summarized as follows.

\noindent \textbf{MomADv2 Framework.}
We propose MomADv2, a reliable temporal memory framework for end-to-end autonomous driving. Unlike prior methods that directly reuse historical planning states, MomADv2 selectively preserves useful temporal information while suppressing stale or command-inconsistent memory. This design improves long-horizon planning consistency, trajectory stability, and robustness under dynamic driving conditions.

\noindent \textbf{SSM-Q and FM-Ref.}
We introduce the Selective State-Space Planning Memory Query Module (SSM-Q), which retrieves historical planning queries based on command consistency, temporal continuity, and trajectory-level alignment, and models the evolution of planning intentions through a selective state-space mechanism. We further propose the Flow-Matching Trajectory Residual Refiner (FM-Ref), which learns a conditional residual velocity field to progressively correct local trajectory deviations while preserving the stability of anchor-based planning.

\subsection{Datasets}
\label{app:datasets_benchmarks}

\noindent\textbf{NAVSIM.}
Following the official NAVSIM protocol, we report results on three
settings: \textit{NAVSIM-v1 navtest}, \textit{NAVSIM-v2 navtest},
and \textit{NAVSIM-v2 navhard}.
The \textit{navtest} split evaluates general planning ability on
standard real-world driving scenes, while \textit{navhard} focuses
on more challenging and safety-critical long-tail scenarios.
\textit{NAVSIM-v1} uses the Predictive Driver Model Score (PDMS)
to assess key aspects of driving behavior, including safety,
feasibility, comfort, and progress.
\textit{NAVSIM-v2} adopts the Extended Predictive Driver Model
Score (EPDMS), which further incorporates rule- and comfort-related
metrics such as driving-direction compliance, traffic-light
compliance, lane keeping, and extended comfort.
For \textit{NAVSIM-v2 navhard}, we follow the official two-stage
protocol: Stage 1 evaluates the planner on the original observation,
and Stage 2 re-evaluates it with synthesized future observations
around the Stage-1 endpoint.
All NAVSIM scores are reported as percentages, and higher values
indicate better planning quality.

\noindent\textbf{Bench2Drive.}
We further evaluate closed-loop driving performance on
Bench2Drive~\cite{jia2024bench2drive}, a CARLA-based benchmark designed
for fine-grained multi-ability evaluation.
Its evaluation set contains 220 short routes covering 44 interactive
scenarios across diverse towns and weather conditions.
Unlike open-loop log-replay evaluation, the executed actions affect
subsequent vehicle states and sensor observations, enabling direct
assessment of planning performance under closed-loop interactions.
Following the official protocol, we report Driving Score (DS),
Success Rate (SR), Efficiency (Effi.), Comfortness (Comf.), and five
advanced driving abilities: Merging, Overtaking, Emergency Brake,
Give Way, and Traffic Sign.
For open-loop evaluation on the logged Bench2Drive data, we
additionally report the average L2 displacement error and trajectory
diversity.

\noindent\textbf{nuScenes.}
We additionally conduct open-loop planning experiments on the
nuScenes validation set~\cite{nuscenes}.
nuScenes contains 1,000 real-world driving scenes collected using
a full surround-view sensor suite, with each scene lasting
approximately 20 seconds.
Under the open-loop protocol, the planner predicts future ego
trajectories from logged observations without affecting subsequent
scene evolution.
We evaluate short-horizon planning at 1, 2, and 3 seconds and
long-horizon planning at 4, 5, and 6 seconds.
The reported metrics include L2 displacement error, collision rate,
and, for multi-mode planners, trajectory diversity.
Together, NAVSIM, Bench2Drive, and nuScenes provide complementary
evaluations of pseudo-closed-loop planning quality, closed-loop
interaction ability, open-loop accuracy, safety, multimodality,
and long-horizon robustness.


\subsection{Evaluation Metrics}
\label{app:evaluation_metrics}


\textbf{PDM and PDMS}.
Following NAVSIM~\cite{navsimv1}, the Predictive Driver Model (PDM)
refers to the rule-based planner used to generate and score trajectory
proposals.
Given an observation $o_t$ and a set of candidate trajectories
$\mathcal{T}=\{\tau_i\}_{i=1}^{N}$, PDM selects the trajectory with
the highest PDM score:
\begin{equation}
\tau^{\star}
=
\arg\max_{\tau_i\in\mathcal{T}}
\mathrm{Score}_{\mathrm{PDM}}(\tau_i),
\label{eq:appendix_pdm_selection}
\end{equation}
where $\mathrm{Score}_{\mathrm{PDM}}$ evaluates each candidate by
simulating it and aggregating safety, feasibility, progress, and
comfort terms.
NAVSIM adopts this PDM-style scoring function as the benchmark
metric, namely the Predictive Driver Model Score (PDMS).

For \textit{NAVSIM-v1}, evaluation first unrolls the predicted
trajectory in a non-reactive simulator and computes normalized
subscores in $[0,1]$.
These subscores are divided into hard penalties and soft objectives.
The hard penalties include no-at-fault collision (NC) and
drivable-area compliance (DAC), while the soft objectives include
ego progress (EP), time-to-collision (TTC), and comfort (C).
Let
$\mathcal{M}_{\mathrm{hard}}=\{\mathrm{NC},\mathrm{DAC}\}$
and
$\mathcal{M}_{\mathrm{soft}}=\{\mathrm{EP},\mathrm{TTC},\mathrm{C}\}$.
The final score is computed as
\begin{equation}
\begin{aligned}
\mathrm{PDMS}
&=
\underbrace{
\prod_{m\in\mathcal{M}_{\mathrm{hard}}}s_m
}_{\text{hard penalties}}
\\[-0.2em]
&\quad\times
\underbrace{
\frac{
\sum_{m\in\mathcal{M}_{\mathrm{soft}}}
\alpha_m s_m
}{
\sum_{m\in\mathcal{M}_{\mathrm{soft}}}\alpha_m
}
}_{\substack{\text{weighted soft}\\\text{score}}}.
\end{aligned}
\label{eq:appendix_pdms}
\end{equation}
Here, $s_m$ denotes the normalized subscore.
NAVSIM uses
$\alpha_{\mathrm{EP}}=5$,
$\alpha_{\mathrm{TTC}}=5$, and
$\alpha_{\mathrm{C}}=2$.
The multiplicative penalty term makes safety-critical violations
dominate the final score: if the ego trajectory causes an at-fault
collision, $s_{\mathrm{NC}}$ becomes zero; if the trajectory leaves
the drivable area, $s_{\mathrm{DAC}}$ becomes zero.
Collisions with static objects are assigned a softer penalty in
NAVSIM, while non-at-fault collisions under the non-reactive setting
are ignored.

The soft terms measure driving quality when the trajectory is
admissible.
The ego-progress subscore $s_{\mathrm{EP}}$ is computed as the ratio
between the ego progress along the route centerline and a safe
upper-bound progress estimated by the privileged PDM-Closed planner,
clipped to $[0,1]$.
The time-to-collision subscore $s_{\mathrm{TTC}}$ is initialized to
one and is set to zero if, at any simulation step within the
4-second horizon, the projected ego motion violates the predefined
TTC safety threshold with respect to surrounding vehicles.
The comfort subscore $s_{\mathrm{C}}$ evaluates whether the trajectory
satisfies acceleration and jerk thresholds.
Therefore, PDMS rewards trajectories that simultaneously remain
collision-free, stay on the road, make sufficient route progress,
preserve safety margins, and maintain smooth motion.


\textbf{EPDMS}.
For \textit{NAVSIM-v2}, the benchmark extends PDMS to the Extended
Predictive Driver Model Score (EPDMS) by adding more fine-grained
rule-compliance and comfort terms.
In addition to NC, DAC, EP, and TTC, EPDMS includes
driving-direction compliance (DDC), traffic-light compliance (TLC),
lane keeping (LK), history comfort (HC), and extended comfort (EC).
Following the NAVSIM-v2 protocol~\cite{navsimv2}, the single-stage
extended score can be written as
\begin{equation}
\begin{aligned}
\mathrm{EPDMS}
&=
\underbrace{
\prod_{m\in\mathcal{M}_{\mathrm{pen}}}
f_m(\tau_{\mathrm{agent}},\tau_{\mathrm{human}})
}_{\text{penalty terms}}
\\[-0.2em]
&\quad\times
\underbrace{
\frac{
\sum_{m\in\mathcal{M}_{\mathrm{avg}}}
\beta_m
f_m(\tau_{\mathrm{agent}},\tau_{\mathrm{human}})
}{
\sum_{m\in\mathcal{M}_{\mathrm{avg}}}\beta_m
}
}_{\substack{\text{weighted average}\\\text{terms}}}.
\end{aligned}
\label{eq:appendix_epdms}
\end{equation}
Here,
\begin{equation}
\begin{aligned}
\mathcal{M}_{\mathrm{pen}}
&=
\{\mathrm{NC},\mathrm{DAC},\mathrm{DDC},\mathrm{TLC}\},\\
\mathcal{M}_{\mathrm{avg}}
&=
\{\mathrm{EP},\mathrm{TTC},\mathrm{LK},\mathrm{HC},\mathrm{EC}\}.
\end{aligned}
\label{eq:appendix_epdms_sets}
\end{equation}
The default weights are
$\beta_{\mathrm{EP}}=5$,
$\beta_{\mathrm{TTC}}=5$, and
$\beta_{\mathrm{LK}}
=\beta_{\mathrm{HC}}
=\beta_{\mathrm{EC}}=2$.

The function
$f_m(\tau_{\mathrm{agent}},\tau_{\mathrm{human}})$
denotes the human-filtered subscore for metric $m$.
This filtering mechanism ignores a rule violation when the same
violation is also committed by the human trajectory in the
corresponding scene, reducing false penalties caused by annotation
noise or contextually necessary maneuvers.
DDC evaluates whether the ego vehicle follows the legal driving
direction, TLC checks compliance with traffic lights, LK evaluates
lane-keeping behavior, and HC and EC measure trajectory smoothness
under the extended NAVSIM-v2 protocol.


\textbf{NAVSIM-v2 Navhard}.
For \textit{NAVSIM-v2 navhard}, evaluation follows a two-stage
pseudo-simulation protocol.
Stage 1 scores the planner from the original real observation,
producing $s_1$.
Stage 2 evaluates the planner on a set of pre-generated synthetic
observations around plausible future ego states, producing
$\{s_2^{(i)}\}_{i=1}^{K}$.
These Stage-2 scores are aggregated by a Gaussian-weighted average
according to the distance between each synthetic start point $x_i$
and the Stage-1 endpoint $\hat{x}$:
\begin{equation}
\begin{aligned}
s_2
&=
\sum_{i=1}^{K}
\hat{w}_i s_2^{(i)},\\
\hat{w}_i
&=
\frac{w_i}{\sum_{j=1}^{K}w_j},\\
w_i
&=
\exp\left(
-\frac{
\lVert x_i-\hat{x}\rVert_2^2
}{
2\sigma^2
}
\right).
\end{aligned}
\label{eq:appendix_stage2_weighting}
\end{equation}
The final navhard score multiplies the original-observation score
and the aggregated synthetic-observation score:
\begin{equation}
\mathrm{EPDMS}_{\mathrm{navhard}}
=
s_1 s_2.
\label{eq:appendix_navhard_epdms}
\end{equation}
This two-stage aggregation evaluates both immediate planning quality
and robustness to future observation shifts, making navhard stricter
than the single-stage NAVSIM-v1 PDMS or NAVSIM-v2 navtest evaluation.


\textbf{Bench2Drive Metrics}.
Bench2Drive~\cite{jia2024bench2drive} evaluates end-to-end driving under
closed-loop interactions.
Success Rate (SR) measures the proportion of routes completed within
the allotted time without traffic violations.
Driving Score (DS) jointly considers route completion and infraction
penalties:
\begin{equation}
\begin{aligned}
\mathrm{SR}
&=
\frac{
N_{\mathrm{succ}}
}{
N_{\mathrm{route}}
},\\
\mathrm{DS}
&=
\frac{1}{N_{\mathrm{route}}}
\sum_{r=1}^{N_{\mathrm{route}}}
\mathrm{RC}_{r}
\prod_{q=1}^{Q_r}p_{r,q}.
\end{aligned}
\label{eq:appendix_bench2drive}
\end{equation}
Here, $N_{\mathrm{succ}}$ and $N_{\mathrm{route}}$ denote the
numbers of successfully completed and total routes, respectively.
$\mathrm{RC}_{r}$ is the route-completion percentage of route $r$,
$Q_r$ is the number of infractions occurring on that route, and
$p_{r,q}$ is the penalty associated with its $q$-th infraction.
Consequently, DS gives partial credit for route progress while
penalizing collisions, traffic-rule violations, and other unsafe
behaviors.

Efficiency measures the average ratio between the ego-vehicle speed
and the average speed of nearby vehicles at regularly spaced route
checkpoints.
It may exceed $100\%$ when the ego vehicle travels faster than the
surrounding traffic.
Comfortness measures the proportion of trajectory segments satisfying
predefined bounds on longitudinal and lateral acceleration, yaw rate,
yaw acceleration, and jerk.

Bench2Drive further groups its interactive scenarios into five
advanced abilities: Merging, Overtaking, Emergency Brake, Give Way,
and Traffic Sign.
The corresponding ability-wise scores provide a more fine-grained
assessment than the overall DS and SR, while their arithmetic mean is
reported as the overall Multi-Ability score.
Higher values indicate better performance for all closed-loop
Bench2Drive metrics.
For open-loop Bench2Drive evaluation, we additionally report the
average L2 displacement error and average trajectory diversity,
using the same definitions as the nuScenes metrics below.


\textbf{nuScenes Metrics}.
For open-loop planning on nuScenes~\cite{nuscenes}, we evaluate the
selected ego trajectory using L2 displacement error and collision
rate.
Given $N$ evaluation samples, the displacement error at future
timestamp $t$ is
\begin{equation}
\mathrm{L2}^{(t)}
=
\frac{1}{N}
\sum_{n=1}^{N}
\left\|
\hat{\mathbf{p}}_{n}^{(t)}
-
\mathbf{p}_{n,\mathrm{gt}}^{(t)}
\right\|_2,
\label{eq:appendix_nuscenes_l2}
\end{equation}
where $\hat{\mathbf{p}}_{n}^{(t)}$ and
$\mathbf{p}_{n,\mathrm{gt}}^{(t)}$ denote the predicted and
ground-truth ego positions of sample $n$ at timestamp $t$,
respectively.
The average L2 error is obtained by averaging
$\mathrm{L2}^{(t)}$ over the reported future timestamps.
Lower values indicate more accurate imitation of the logged human
trajectory.

Following SparseDrive~\cite{sun2024sparsedrive}, collision rate measures the
percentage of evaluation samples in which the predicted oriented ego
footprint overlaps with any surrounding actor at the evaluated
horizon.
The ego heading is estimated from consecutive trajectory points
instead of being assumed constant.
This oriented-box implementation reduces false collision detections
caused by coarse occupancy rasterization or fixed-heading
approximations.
A lower collision rate indicates safer planning.

For a multi-mode planner producing $M$ candidate trajectories, we
additionally adopt the time-conditioned trajectory Diversity Metric
$\mathrm{Div}^{(t)}$~\cite{diver}.
Let $\mathbf{p}_{t}^{(i)}$ denote the waypoint of the $i$-th
trajectory mode at timestamp $t$.
The metric is computed as
\begin{equation}
\begin{aligned}
d_{ij}^{(t)}
&=
\left\|
\mathbf{p}_{t}^{(i)}
-
\mathbf{p}_{t}^{(j)}
\right\|_2,\\
D_{\mathrm{raw}}^{(t)}
&=
\frac{2}{M(M-1)}
\sum_{i=1}^{M-1}
\sum_{j=i+1}^{M}
d_{ij}^{(t)},\\
Z_t
&=
\epsilon
+
\frac{1}{M}
\sum_{i=1}^{M}
\left\|
\mathbf{p}_{t}^{(i)}
\right\|_2,\\
\mathrm{Div}^{(t)}
&=
\min\left(
1,
\frac{
D_{\mathrm{raw}}^{(t)}
}{
Z_t
}
\right).
\end{aligned}
\label{eq:appendix_diversity}
\end{equation}
Here, $D_{\mathrm{raw}}^{(t)}$ is the mean pairwise distance between
trajectory modes, $Z_t$ normalizes this distance according to the
trajectory scale, and $\epsilon$ is a small constant preventing
division by zero.
The resulting metric is bounded within $[0,1]$.
A higher value indicates that the predicted modes are more diverse
and better separated in trajectory space, whereas a value close to
zero indicates mode collapse.
The average diversity is computed by averaging
$\mathrm{Div}^{(t)}$ over the reported future timestamps.
This metric is applicable only to multi-mode planning methods.

\subsection{More Details of MomADv2}

\noindent\textbf{Temporal-Link Construction and Memory Reset.}
The temporal-continuity token used by SSM-Q is a frame-level identifier
rather than a learnable representation. Let
$\operatorname{prev}(\chi_t)$ denote the identifier of the immediate
predecessor of frame $t$ according to the sequential metadata. We define
\begin{equation}
\chi_t^{(-k)}
=
\operatorname{prev}^{k}(\chi_t),
\end{equation}
where $\operatorname{prev}^{k}$ applies the predecessor operation $k$
times. Therefore, a historical entry at time $t-k$ is temporally valid only
when $\chi_t^{(-k)}=\chi_{t-k}$. The identifier is used only for memory
indexing and hard validity masking; it is not embedded or provided as an
input feature to the SSM encoder.

The memory bank is reset at the beginning of each sequence and whenever the
current frame belongs to a new scene, has no valid predecessor, or violates
the expected frame interval. During the first $K$ frames of a sequence, only
the available valid entries are used. If no valid historical entry exists,
the temporal residual is set to zero and the model falls back to the
single-frame planning branch.


\noindent\textbf{Memory Contents and Causal Update.}
At frame $t$, the baseline planner produces the scene-conditioned planning
queries $\mathbf{Q}^{\mathrm{raw}}_t$ and command-specific candidate
trajectories $\mathbf{Y}^{\mathrm{base,cmd}}_t$ before temporal enhancement.
Here, ``raw'' means pre-SSM rather than a fixed planning-anchor embedding.
Each memory entry is represented as
\begin{equation}
\mathcal{E}_t
=
\left(
\chi_t,\,
c_t,\,
\mathbf{Q}^{\mathrm{raw}}_t,\,
\mathbf{Y}^{\mathrm{base,cmd}}_t
\right).
\end{equation}
The implementation indexes each frame-level entry by $\chi_t$ and stores
$c_t$ as command metadata; retrieval therefore treats $(\chi_t,c_t)$ as a
compound logical key. Candidates associated with commands different from
$c_t$ are never eligible for retrieval.

We use a first-in--first-out memory containing at most $K$ historical
entries. Following the memory-length ablation in the main paper, we set
$K=4$ unless otherwise specified. At each frame, the model first reads
$\mathcal{M}_{t-1}$, retrieves and aggregates valid historical states,
produces the final planning result, and only then appends
$\mathcal{E}_t$ to the memory bank. Entries older than $K$ frames are
discarded. Only the pre-enhancement query and baseline trajectory are
stored; neither the SSM-enhanced query nor the flow-refined trajectory is
written back. This read-before-write strategy prevents the current frame
from retrieving itself and avoids recursive amplification of temporal
corrections.


\noindent\textbf{Temporally Shifted Trajectory Distance.}
The trajectory-level alignment in SSM-Q compares historical and current
planning candidates over the same absolute future times. This temporal
alignment is necessary because a trajectory predicted at time $t-k$ starts
$k$ frames earlier than a trajectory predicted at time $t$. Let
$\Delta_{\mathrm{frame}}$ denote the interval between consecutive planning
frames and $\Delta_{\mathrm{traj}}$ denote the interval between trajectory
waypoints. The number of elapsed trajectory steps is
\begin{equation}
s_k
=
\operatorname{round}
\left(
\frac{k\Delta_{\mathrm{frame}}}{\Delta_{\mathrm{traj}}}
\right).
\end{equation}
For a historical candidate $\mathbf{Y}_{t-k,h}$, we first discard its first
$s_k$ waypoints, which correspond to time steps that have already elapsed.
The remaining points are transformed from the historical ego coordinate
system to the current ego coordinate system using the relative ego pose
$\mathcal{T}_{t\leftarrow t-k}$. The shifted trajectory distance is then
defined over the overlapping horizon. Specifically, the aligned historical
waypoint is
\begin{equation}
\widetilde{\mathbf{y}}_{t-k,h}^{\ell}
=
\mathcal{T}_{t\leftarrow t-k}
\left(
\mathbf{y}_{t-k,h}^{\ell+s_k}
\right),
\end{equation}
and the shifted distance is
\begin{equation}
D_{\mathrm{shift}}
=
\frac{\displaystyle
\sum_{\ell=1}^{T-s_k}
\left\|
\widetilde{\mathbf{y}}_{t-k,h}^{\ell}
-
\mathbf{y}_{t,j}^{\ell}
\right\|_2}
{T-s_k}.
\end{equation}
When the temporal offset does not correspond to an integer number of
trajectory steps, the historical trajectory is linearly interpolated at the
timestamps of the current trajectory before computing the distance. Thus,
the second predicted waypoint from the previous frame is compared with the
first predicted waypoint from the current frame when one trajectory step has
elapsed. This avoids treating the same driving intention as inconsistent
solely because the two predictions have different temporal origins.

For each current command-specific candidate $j$, SSM-Q selects the historical
candidate with the minimum shifted distance. The resulting distance is also
used by the similarity weight and reliability gate:
\begin{equation}
d_{t,k,j}
=
D_{\mathrm{shift}}
\left(
\mathbf{Y}_{t-k,h^*},
\mathbf{Y}_{t,j}
\right).
\end{equation}
If $s_k\geq T$, the two trajectories have no overlapping future horizon and
the corresponding historical entry is marked invalid. This operation is
used only for candidate association and reliability estimation: it does not
shift the planning query itself or overwrite the stored trajectory.


\noindent\textbf{Handling Inaccurate Baseline Candidates.}
The baseline trajectory is used as an association cue rather than directly
propagated as the final output. A historical candidate with a large aligned
distance receives a small similarity weight through
$\exp(-d_{t,k,j}/\alpha)$ and a small distance-reliability value through
$r^{\mathrm{dist}}_{t,j}$. Invalid, temporally discontinuous, or
command-inconsistent entries receive zero weight. If all retrieved entries
are rejected, the reliability-gated residual vanishes and the current query
remains unchanged. Consequently, an unreliable baseline match can be
suppressed instead of being unconditionally injected into the current
planning state.

\noindent\textbf{Causal Online Inference.}
MomADv2 performs causal stateful inference without test-time optimization.
For every input frame, the perception backbone is evaluated once. The model
then executes the baseline planning head, historical-memory retrieval,
trajectory-level alignment, SSM-based query enhancement, and the planning
refinement head in sequence. The resulting trajectory is further processed
by the flow refiner using $S=2$ Euler evaluations. Finally, the current
pre-enhancement query and baseline trajectory are appended to the memory
bank for the next frame. All model parameters remain fixed during this
process; only the explicit memory state changes. The proposed inference
procedure is therefore online stateful inference rather than test-time
training or test-time adaptation.


\begin{table*}[t]
\scriptsize
\centering
\addtolength{\tabcolsep}{0.1pt}
\caption{$\operatorname{Multi-Ability}$ results on $\operatorname{\textbf{Bench2Drive}}$ (V0.0.3) under base training set.
`mmt' refers multi-mode trajectory variant of $\operatorname{VAD}$ and $^\dagger$ denotes the re-implementation. * denotes expert feature distillation.
}
\renewcommand\arraystretch{0.7}
\tabcolsep=3.4mm
\resizebox{\linewidth}{!}{
\begin{tabular}{l  cccccc}
\toprule
\multirow{2}{*}{$\operatorname{Method}$}
& \multicolumn{6}{c}{$\operatorname{Multi-Ability (\%) \uparrow}$} \\
\cmidrule(lr){2-7}
&  $\operatorname{Merging}\uparrow$
& $\operatorname{Overtaking}\uparrow$
& $\operatorname{Emergency Brake}\uparrow$
& $\operatorname{Give\:Way}\uparrow$
& $\operatorname{Traffic\:Sign}\uparrow$
& $\operatorname{Mean}\uparrow$ \\

\midrule
DriveAdapter* \cite{jia2023driveadapter} & 28.82 & 26.38 & 48.76 & \textbf{50.00} & 56.43 & 42.08 \\
DriveDPO \cite{shang2025drivedpo}  & - & - & - & - & - & - \\
Raw2Drive \cite{Raw2Drive}  & 43.35 & \textbf{51.11} & \textbf{60.00} & \textbf{50.00} & 62.26 & \textbf{53.34} \\
DriveTrans* \cite{jia2025drivetransformer}  & 17.57 & 35.00 & 48.36 & 40.00 & 52.10 & 38.60 \\
WoTE* \cite{wote}  & - & - & - & - & - & - \\
DIVER \cite{diver}  & 35.08 & 25.09 & 41.09 & \textbf{50.00} & 59.21 & 42.09 \\

\cellcolor{red!5}${\operatorname{MomADv2}}$ (Ours)
& \cellcolor{red!5}\textbf{43.71}
& \cellcolor{red!5}34.39
& \cellcolor{red!5}50.54
& \cellcolor{red!5}\textbf{50.00}
& \cellcolor{red!5}\textbf{69.82}
& \cellcolor{red!5}49.02 \\

\midrule
\midrule
UniAD-Base \cite{uniad} & 14.10 & 17.78 & 21.67 & 10.00 & 14.21 & 15.55 \\
$\operatorname{VAD}$ \cite{jiang2023vad} & 8.11 & 24.44 & 18.64 & \textbf{20.00} & 19.15 & 18.07 \\
GenAD \cite{zheng2024genad} & - & - & - & - & - & - \\
$\operatorname{MomAD}$ \cite{momad} & 13.21 & 21.02 & 18.01 & \textbf{20.00} & 21.07 & 18.66 \\
$\operatorname{DIVER}$ \cite{diver} & 13.83 & 29.09 & 25.51 & \textbf{20.00} & 24.93 & 22.67 \\
\cellcolor{red!5}${\operatorname{MomADv2}}$ (Ours)
& \cellcolor{red!5}\textbf{18.79}
& \cellcolor{red!5}\textbf{32.86}
& \cellcolor{red!5}\textbf{27.73}
& \cellcolor{red!5}\textbf{20.00}
& \cellcolor{red!5}\textbf{27.96}
& \cellcolor{red!5}\textbf{25.84} \\

\bottomrule
\end{tabular}
}
\label{tab_b2d_multi_ability}
\end{table*}

\begin{table}[t]
\centering
\caption{Comparison of trajectory prediction consistency on the
\textbf{nuScenes} validation set. TPC denotes Trajectory Prediction
Consistency, where lower values indicate better temporal stability.}
\renewcommand\arraystretch{0.8}
\tabcolsep=3.2mm
\resizebox{\linewidth}{!}{
\begin{tabular}{l c c c c}
\toprule
Method & Venue & \multicolumn{3}{c}{TPC (m)$\downarrow$} \\
\cmidrule(lr){3-5}
& & 4s & 5s & 6s \\
\midrule
UniAD~\cite{uniad} & CVPR'23 & 1.49 & 1.81 & 2.41 \\
VAD~\cite{jiang2023vad} & ICCV'23 & 1.55 & 1.73 & 2.17 \\
SparseDrive~\cite{sun2024sparsedrive} & ICRA'25 & 1.33 & 1.66 & 1.99 \\
MomAD~\cite{momad} & CVPR'25 & 1.19 & 1.45 & \textbf{1.61} \\
\rowcolor{red!5}
MomADv2 (Ours) & - & \textbf{1.04} & \textbf{1.32} & \textbf{1.61} \\
\bottomrule
\end{tabular}}
\label{tab:tpc_nuscenes}
\end{table}

\begin{table}[t]
\scriptsize
\centering
\caption[]{Planning results for \textbf{3-second short-horizon planning} on the \textbf{nuScenes} validation set.}
\renewcommand\arraystretch{0.7}
\setlength{\tabcolsep}{0.99mm}
\resizebox{\linewidth}{!}{
\begin{tabular}{l cccc cccc}
\toprule
\multirow{2}{*}{$\operatorname{Method}$}
& \multicolumn{4}{c}{$\operatorname{L2\ (m)}\downarrow$}
& \multicolumn{4}{c}{$\operatorname{Col.\ Rate\ (\%)}\downarrow$} \\
\cmidrule(lr){2-5}
\cmidrule(lr){6-9}
& 1s & 2s & 3s & $\operatorname{Avg.}$
& 1s & 2s & 3s & $\operatorname{Avg.}$ \\
\midrule

$\operatorname{UniAD}$~\cite{uniad}
& 0.48 & 0.96 & 1.65 & \cellcolor{red!5}1.03
& 0.10 & 0.15 & 0.61 & \cellcolor{red!5}0.29 \\

$\operatorname{VAD}$~\cite{jiang2023vad}
& 0.41 & 0.70 & 1.05 & \cellcolor{red!5}0.72
& 0.11 & 0.24 & 0.42 & \cellcolor{red!5}0.26 \\

$\operatorname{DiffusionDrive}$~\cite{liao2024diffusiondrive}
& 0.29 & 0.58 & 0.96 & \cellcolor{red!5}0.61
& 0.02 & 0.05 & 0.22 & \cellcolor{red!5}0.09 \\

$\operatorname{DIVER}$~\cite{diver}
& -- & -- & -- & \cellcolor{red!5}--
& 0.01 & 0.05 & \textbf{0.15}
& \cellcolor{red!5}\textbf{0.07} \\

$\operatorname{FocalAD}$~\cite{sun2025focalad}
& 0.27 & 0.57 & 0.96 & \cellcolor{red!5}0.60
& \textbf{0.00} & 0.04 & 0.24
& \cellcolor{red!5}0.09 \\

$\operatorname{GuideFlow}$~\cite{liu2025guideflow}
& -- & -- & -- & \cellcolor{red!5}--
& \textbf{0.00} & \textbf{0.02} & 0.18
& \cellcolor{red!5}\textbf{0.07} \\

$\operatorname{SparseDrive}$~\cite{sun2024sparsedrive}
& 0.30 & 0.58 & 0.96 & \cellcolor{red!5}0.61
& 0.01 & 0.05 & 0.23
& \cellcolor{red!5}0.10 \\

$\operatorname{LAW}$~\cite{li2024enhancing_law}
& 0.26 & 0.57 & 1.01 & \cellcolor{red!5}0.61
& 0.14 & 0.21 & 0.54
& \cellcolor{red!5}0.30 \\

$\operatorname{GenAD}$~\cite{zheng2024genad}
& 0.28 & 0.49 & 0.78 & \cellcolor{red!5}0.52
& 0.08 & 0.14 & 0.34
& \cellcolor{red!5}0.19 \\

$\operatorname{Drive\text{-}OccWorld}$~\cite{zheng2024occworld}
& 0.25 & 0.44 & 0.72 & \cellcolor{red!5}0.47
& 0.03 & 0.08 & 0.22
& \cellcolor{red!5}0.11 \\

$\operatorname{SSR}$~\cite{ssr}
& \textbf{0.19} & \textbf{0.36} & \textbf{0.62}
& \cellcolor{red!5}\textbf{0.39}
& 0.10 & 0.10 & 0.24
& \cellcolor{red!5}0.15 \\

$\operatorname{MomAD}$~\cite{momad}
& 0.31 & 0.57 & 0.91 & \cellcolor{red!5}0.60
& 0.01 & 0.05 & 0.22
& \cellcolor{red!5}0.09 \\

$\operatorname{GraphWorld}$~\cite{song2026graphworld}
& 0.29 & 0.55 & 0.89 & \cellcolor{red!5}0.57
& \textbf{0.00} & 0.04 & 0.20
& \cellcolor{red!5}0.08 \\

\cellcolor{red!5}$\operatorname{MomADv2(Ours)}$
& \cellcolor{red!5}0.26
& \cellcolor{red!5}0.52
& \cellcolor{red!5}0.85
& \cellcolor{red!5}0.54
& \cellcolor{red!5}\textbf{0.00}
& \cellcolor{red!5}0.03
& \cellcolor{red!5}0.18
& \cellcolor{red!5}\textbf{0.07} \\

\bottomrule
\end{tabular}
}
\label{tab_nuscenes_planning_3s}
\end{table}

\begin{table}[t]
\centering
\caption{Ablation study of the two core modules on the
\textbf{nuScenes} validation set under open-loop evaluation.}
\renewcommand\arraystretch{0.8}
\tabcolsep=3.2mm
\resizebox{\linewidth}{!}{
\begin{tabular}{c c c c c c c c}
\toprule
\multicolumn{2}{c}{} &
\multicolumn{3}{c}{$L_2$ (m)$\downarrow$} &
\multicolumn{3}{c}{Col. Rate (\%)$\downarrow$} \\
\cmidrule(lr){3-5}
\cmidrule(lr){6-8}
SSM-Q & FM-Ref
& 4s & 5s & 6s
& 4s & 5s & 6s \\
\midrule
&
& 1.67 & 1.98 & 2.45
& 0.87 & 1.54 & 2.33 \\

$\checkmark$ &
& 1.33 & 1.84 & 2.43
& 0.80 & 1.44 & 2.12 \\

\rowcolor{red!5}
$\checkmark$ & $\checkmark$
& \textbf{1.32}
& \textbf{1.82}
& \textbf{2.40}
& \textbf{0.72}
& \textbf{1.33}
& \textbf{2.03} \\
\bottomrule
\end{tabular}}
\label{tab:ablation_nus}
\end{table}

\begin{table}[t]
\centering
\caption{\textcolor{black}{Ablation study of temporal memory modeling methods. 
Avg. L2 and Avg. Col. are evaluated on \textbf{nuScenes}, and EPDMS is evaluated on \textbf{NAVSIMv2 navtest}.}}
\renewcommand\arraystretch{0.8}
\tabcolsep=2.2mm
\resizebox{\linewidth}{!}{
\begin{tabular}{l c c c c}
\toprule
Temporal Modeling & Avg. L2$\downarrow$ & Avg. Col.$\downarrow$ & EPDMS$\uparrow$ & FPS$\uparrow$ \\
\midrule
Concat + MLP 
& 1.52 & 1.26 & 84.2 & \textbf{51.0} \\

GRU 
& 1.41 & 0.77 & 86.4 & 41.9 \\

LSTM 
& 1.34 & 0.80 & 86.7 & 39.6 \\

Transformer 
& 1.31 & 0.82 & 87.2 & 33.0 \\

Vanilla SSM 
& 1.25 & 0.81 & 87.0 & 43.6 \\

\rowcolor{red!5}
Selective SSM / Mamba 
& \textbf{1.21} & \textbf{0.76} & \textbf{87.9} & 43.0 \\
\bottomrule
\end{tabular}}
\label{tab_ablation_temporal_modeling}
\end{table}

\begin{table}[!htp]
\scriptsize
\centering
\caption{Ablation on solver choice on \textbf{NAVSIMv2 navtest}.}
\renewcommand\arraystretch{0.7}
\setlength{\tabcolsep}{2.0mm}
\resizebox{\linewidth}{!}{
\begin{tabular}{l cccccc c}
\toprule
Solver & NC$\uparrow$ & DAC$\uparrow$ & TTC$\uparrow$ & Comf.$\uparrow$ & EP$\uparrow$ & PDMS$\uparrow$ & FPS$\uparrow$ \\
\midrule
\rowcolor{red!5}Euler& 99.0& 97.1& 95.5& 100& 83.2& 89.9& 43 \\
Heun  & 98.5 & 96.7 & 95.2 & 100 & 82.9 & 89.0 & 38 \\
\bottomrule
\end{tabular}}
\label{tab_solver_choice}
\end{table}

\begin{figure*}[t]
  \centering
  \includegraphics[width=\linewidth]{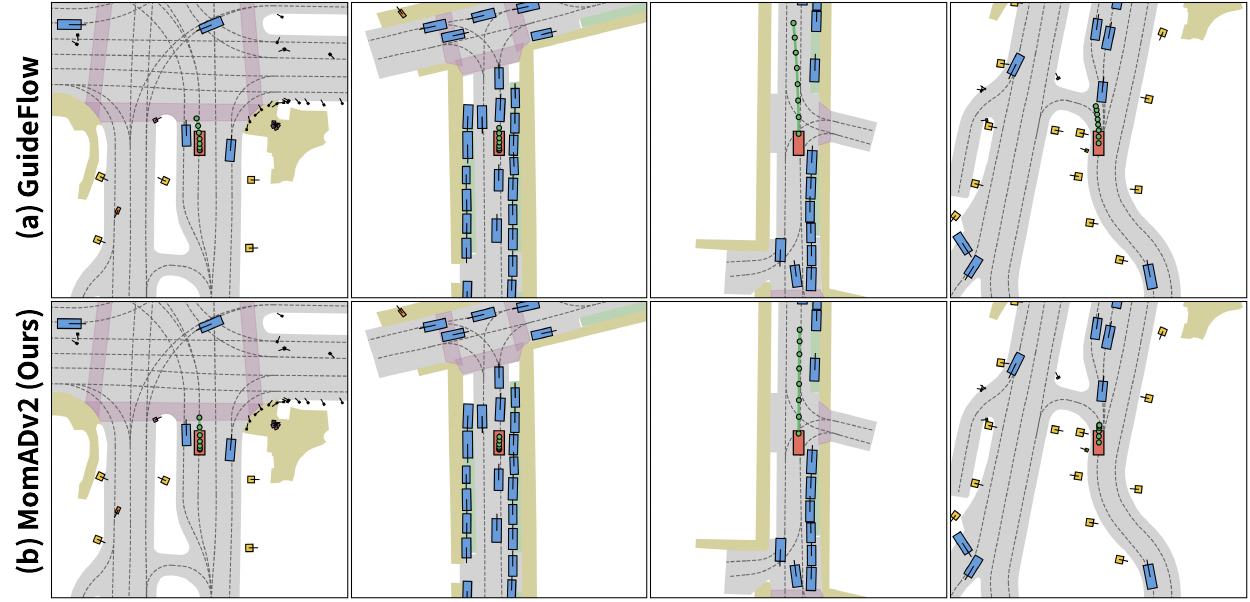}
  \caption{\textbf{Qualitative visualization of MomADv2 on NAVSIM across diverse driving scenarios.}
Compared with GuideFlow, MomADv2 generates smoother and more temporally consistent trajectories, better adheres to lane geometry and drivable-area constraints, and enables safer interactions with surrounding traffic participants.}
  \label{Navsim_vis_fl}
\end{figure*}

\begin{figure*}[t]
  \centering
  \includegraphics[width=\linewidth]{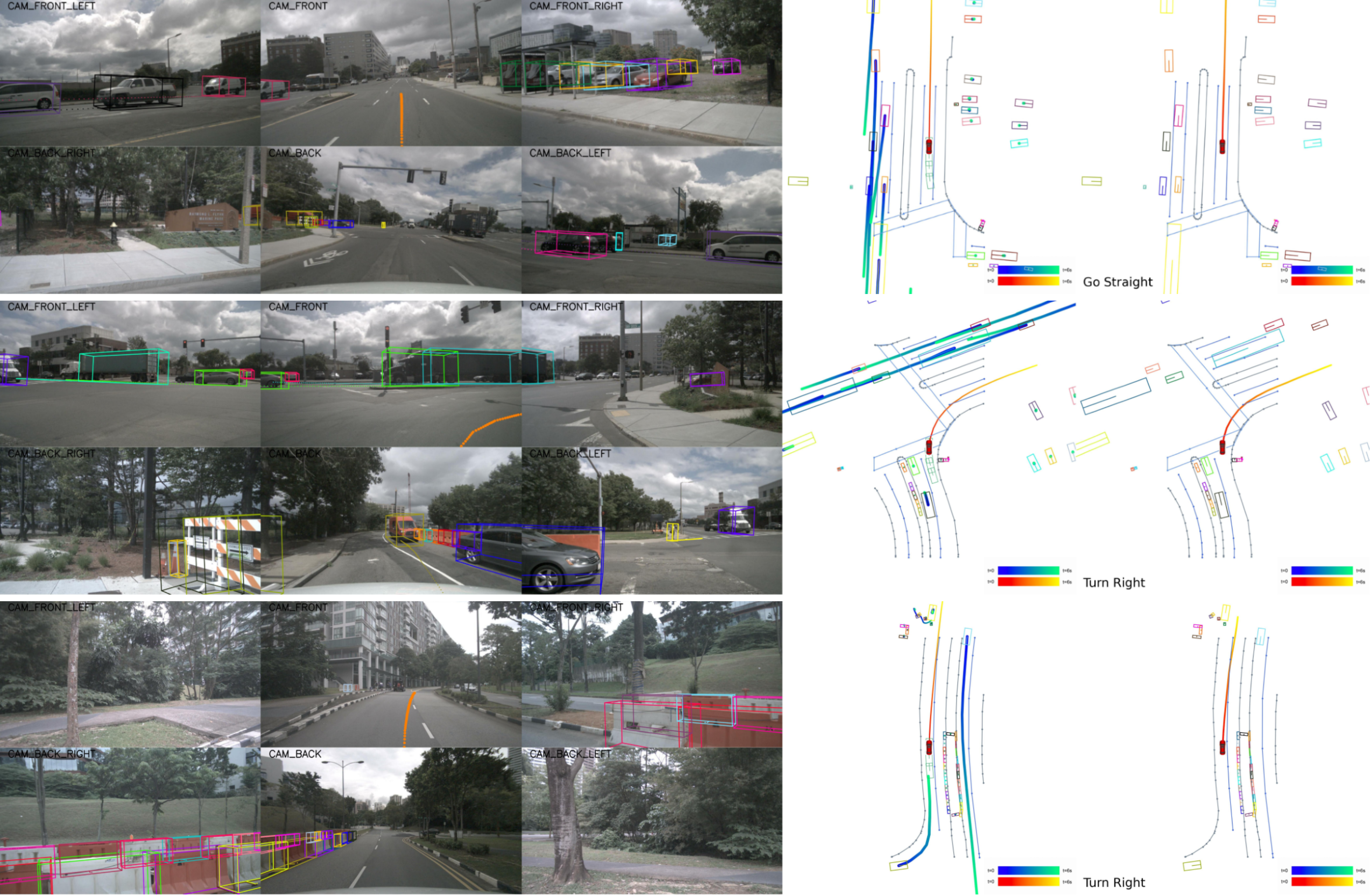}
  \caption{\textbf{Qualitative visualization of 6-second long-horizon planning by MomADv2 on nuScenes.}
}
  \label{nus_vis_fl}
\end{figure*}

\subsection{Implementation Details}

\noindent\textbf{Training Protocol on nuScenes.}
For the nuScenes experiments, we adopt MomAD as the baseline and train MomADv2 using a two-stage optimization protocol.

\noindent\textit{Stage I: Baseline Pretraining.}
We first follow the original MomAD training procedure to obtain a fully trained end-to-end autonomous driving model comprising sparse perception, motion prediction, and ego planning modules. Based on this pretrained model, most components of the original Topological Trajectory Matching (TTM) and Momentum Planning Interactor (MPI) modules are replaced with the proposed Selective State-Space Planning Memory Query Module (SSM-Q) and Flow-Matching Trajectory Residual Refiner (FM-Ref).

\noindent\textit{Stage II: Module Adaptation.}
MomADv2 is initialized from the pretrained MomAD checkpoint. During this stage, the image backbone, sparse perception modules, motion prediction branch, and most parameters of the original planning head are frozen. We optimize only the newly introduced temporal memory and trajectory refinement components, including the selective state-space encoder, the SSM-related reliability and residual gates, the trajectory residual prediction head, and the flow-matching refinement head. When enabled, the planning regression branch in the final refinement layer is also jointly fine-tuned to improve the integration of the proposed modules with the original anchor-based planner. All Batch Normalization layers are maintained in evaluation mode to preserve the pretrained feature statistics.

\noindent\textbf{Training Configuration.}
The second stage is trained for 10 epochs with a learning rate of $3 \times 10^{-6}$. Following the baseline configuration, we employ ResNet-50 as the backbone with an input resolution of $256 \times 704$. The detection range is defined as a circular region with a radius of 55~m, while the online mapping range is set to $60 \times 30$~m along the longitudinal and lateral directions, respectively. The motion prediction branch maintains six trajectory modes. All experiments are conducted on 8 NVIDIA RTX 4090 GPUs.

\noindent\textbf{Optimization Rationale.}
The first stage establishes a strong and stable perception--motion--planning baseline, whereas the second stage focuses on adapting SSM-Q and FM-Ref without disrupting the pretrained representation and planning capabilities. This decoupled optimization strategy mitigates interference caused by large-scale parameter updates, stabilizes the integration of reliable temporal memory and residual trajectory refinement, and improves training efficiency under a limited additional optimization budget.

\subsection{More Planning Results}
\noindent \textbf{Multi-Ability Evaluation on Bench2Drive.}
As shown in Table~\ref{tab_b2d_multi_ability}, we further evaluate the driving ability of MomADv2 under diverse interactive scenarios.
With expert feature distillation, MomADv2 achieves the best performance on Merging and Traffic Sign, and obtains competitive mean performance of 49.02\%, outperforming DriveAdapter and DIVER by 16.5\% and 16.5\%, respectively.
Without expert feature distillation, MomADv2 consistently ranks first across all abilities, improving the mean score from 22.67\% to 25.84\% compared with DIVER.
These results demonstrate that MomADv2 improves complex driving behaviors such as merging, overtaking, emergency braking, and traffic-rule compliance.

\noindent \textbf{Trajectory Prediction Consistency in nuScenes.} As shown in Table~\ref{tab:tpc_nuscenes}, we further evaluate temporal planning consistency on the nuScenes validation set using the Trajectory Prediction Consistency (TPC) metric introduced by MomAD. MomADv2 achieves the lowest TPC errors of 1.04~m and 1.32~m at the 4- and 5-second horizons, corresponding to relative improvements of 12.6\% and 9.0\% over MomAD, respectively. At the more challenging 6-second horizon, MomADv2 obtains a TPC error of 1.61~m, matching the best-performing baseline. Overall, MomADv2 reduces the average TPC error by 6.6\% compared with MomAD, demonstrating that selective temporal memory and flow-matching-based trajectory refinement improve temporal stability and mitigate planning drift over extended horizons.

\noindent \textbf{3-second short-horizon planning on nuScenes.}
Table~\ref{tab_nuscenes_planning_3s} reports the short-horizon planning results on the nuScenes validation set. MomADv2 achieves an average collision rate of 0.07\%, matching the best performance of DIVER and GuideFlow, while maintaining zero collisions at the 1-second horizon. Compared with MomAD, MomADv2 reduces the average collision rate from 0.09\% to 0.07\% and the 3-second collision rate from 0.22\% to 0.18\%. It also improves over GraphWorld, reducing its average collision rate from 0.08\% to 0.07\%. In terms of trajectory accuracy, MomADv2 obtains an average $L_2$ error of 0.54\,m, improving upon MomAD and GraphWorld by 10.0\% and 5.3\%, respectively. Although SSR achieves the lowest displacement error, its average collision rate is notably higher than that of MomADv2. These results demonstrate that MomADv2 maintains competitive short-horizon trajectory accuracy while providing a favorable balance between planning accuracy and safety.

\subsection{More Ablation Results}
\noindent\textbf{Roles of Different Modules in MomADv2.}
As shown in Table~\ref{tab:ablation_nus}, we further evaluate the contributions of SSM-Q and FM-Ref on the nuScenes validation set under open-loop evaluation. Introducing SSM-Q consistently reduces both trajectory error and collision rate across all planning horizons, demonstrating the effectiveness of reliable temporal memory for long-horizon planning. Further incorporating FM-Ref yields additional improvements, achieving the lowest $L_2$ errors of 1.32, 1.82, and 2.40~m, together with collision rates of 0.72\%, 1.33\%, and 2.03\% at 4, 5, and 6 seconds, respectively. These results confirm that selective temporal memory and flow-matching residual refinement provide complementary benefits for accurate and safe long-horizon planning.

\noindent\textbf{Effectiveness of Selective State-Space Memory Modeling.}
Table~\ref{tab_ablation_temporal_modeling} compares different temporal memory modeling methods.
Concat + MLP is efficient but lacks temporal modeling ability, while GRU/LSTM and Transformer improve planning performance at the cost of higher latency.
Vanilla SSM achieves a better efficiency-performance trade-off, and Selective SSM/Mamba further obtains the best planning accuracy and collision performance.
This shows that selective state-space modeling is more suitable for reliable long-horizon planning memory.

\noindent\textbf{Effect of Solver Choice.}
Table~\ref{tab_solver_choice} compares different solvers in the Flow-Matching Trajectory Residual Refiner.
Euler achieves better planning performance than Heun, improving PDMS from 89.0 to 89.9 while also increasing FPS from 38 to 43.
This shows that the simple Euler solver is sufficient for residual trajectory refinement and provides a better trade-off between planning quality and inference efficiency.

\subsection{More Qualitative Study of Planning Results}

To further examine the planning behavior of MomADv2, we conduct qualitative studies on both NAVSIM and nuScenes. We select representative scenarios involving intersections, curved roads, dense traffic, and long-horizon maneuvering. Specifically, we provide two types of visualization: Visualization on NAVSIM, and Long-Horzion Planning Visualization on nuScenes.

\noindent\textbf{Visualization on NAVSIM.}
We qualitatively compare MomADv2 with GuideFlow across diverse driving scenarios. As illustrated in Fig.~\ref{Navsim_vis_fl}, MomADv2 generates smoother and more temporally consistent trajectories, while better conforming to lane geometry and drivable-area constraints. In complex intersections and curved-road scenarios, MomADv2 also maintains safer interactions with surrounding traffic participants and avoids overly aggressive trajectory deviations. These results demonstrate that reliable temporal memory improves planning continuity and closed-loop decision stability.

\noindent\textbf{Long-Horzion Planning Visualization on nuScenes.}
We further qualitatively examine the 6-second planning performance of MomADv2 on the nuScenes validation set. As illustrated in Fig.~\ref{nus_vis_fl}, MomADv2 generates smooth and temporally coherent trajectories across diverse driving scenarios, including turning maneuvers and dense traffic. By selectively retaining reliable historical context and suppressing stale or command-inconsistent memory, MomADv2 maintains stable planning intentions over extended horizons. The flow-matching trajectory residual refiner further mitigates local deviations and accumulated errors, leading to more accurate and consistent long-horizon planning.

\end{document}